\documentclass[letterpaper, 10pt, conference]{ieeeconf}
\IEEEoverridecommandlockouts
\usepackage{cite}
\usepackage{amsmath,amssymb,amsfonts}
\usepackage{algorithmic}
\usepackage{graphicx}
\usepackage{textcomp}
\usepackage[table]{xcolor}
\usepackage{tikz}
\usepackage{makecell}
\usetikzlibrary{arrows.meta}
\usetikzlibrary{plotmarks} %
\def\BibTeX{{\rm B\kern-.05em{\sc i\kern-.025em b}\kern-.08em
    T\kern-.1667em\lower.7ex\hbox{E}\kern-.125emX}}

\usepackage{booktabs}
\usepackage{multirow}
\usepackage{capt-of}   %

\definecolor{blueFill}{HTML}{EAF2FD}
\definecolor{blueStroke}{HTML}{1F6FEB}
\definecolor{blueStrongFill}{HTML}{DBEAFE}
\definecolor{purpleFill}{HTML}{F2ECFD}
\definecolor{purpleStroke}{HTML}{ff0092}
\definecolor{purpleStrongFill}{HTML}{ffdcf0}
\definecolor{amberFill}{HTML}{FDF1E0}
\definecolor{amberStroke}{HTML}{D97706}
\definecolor{amberStrongFill}{HTML}{FEF3C7}
\definecolor{greenFill}{HTML}{E6F7EF}
\definecolor{greenStroke}{HTML}{059669}
\definecolor{greenStrongFill}{HTML}{D1FAE5}
\definecolor{slateFill}{HTML}{EEF1F5}
\definecolor{slateStroke}{HTML}{475569}
\definecolor{panelTrainStroke}{HTML}{94A3B8}
\definecolor{panelDeployStroke}{HTML}{10B981}
\definecolor{textDark}{HTML}{0F172A}
\definecolor{textMed}{HTML}{334155}
\definecolor{ctPush}{HTML}{CDE2FB}
\definecolor{ctCarry}{HTML}{9EC5F4}
\definecolor{ctGrasp}{HTML}{6DA7EC}
\definecolor{grpBand}{HTML}{F0EFEC}

\tikzset{
  every node/.style={font=\sffamily},
  boxblue/.style={draw=blueStroke, fill=blueFill, line width=0.8pt, rounded corners=1.8pt},
  boxbluestrong/.style={draw=blueStroke, fill=blueStrongFill, line width=1.3pt, rounded corners=1.8pt},
  boxpurple/.style={draw=purpleStroke, fill=purpleFill, line width=0.8pt, rounded corners=1.8pt},
  boxpurplestrong/.style={draw=purpleStroke, fill=purpleStrongFill, line width=1.3pt, rounded corners=1.8pt},
  boxamber/.style={draw=amberStroke, fill=amberFill, line width=0.8pt, rounded corners=1.8pt},
  boxamberstrong/.style={draw=amberStroke, fill=amberStrongFill, line width=1.3pt, rounded corners=1.8pt},
  boxgreenstrong/.style={draw=greenStroke, fill=greenStrongFill, line width=1.3pt, rounded corners=1.8pt},
  boxslate/.style={draw=slateStroke, fill=slateFill, line width=0.8pt, rounded corners=1.8pt},
  container/.style={draw=gray!55, fill=white, line width=0.8pt, rounded corners=2.8pt},
  panel/.style={draw=panelTrainStroke, fill=gray!2, line width=1.1pt, rounded corners=4pt, dash pattern=on 5pt off 3pt},
  paneldeploy/.style={draw=greenStroke, fill=green!2, line width=1.1pt, rounded corners=4pt, dash pattern=on 5pt off 3pt},
  pill/.style={draw=greenStroke, fill=white, line width=0.9pt, rounded corners=0.28cm},
  labelbg/.style={draw=amberStroke, fill=amberStrongFill, rounded corners=1.5pt, inner sep=2pt, font=\tiny},
  labelbgblue/.style={draw=blueStroke, fill=blueStrongFill, rounded corners=1.5pt, inner sep=2pt, font=\tiny},
  flow/.style={-{Stealth[length=2.2mm]}, draw=slateStroke, line width=0.9pt},
  flowblue/.style={-{Stealth[length=2.2mm]}, draw=blueStroke, line width=1.2pt},
  flowpurple/.style={-{Stealth[length=2.2mm]}, draw=purpleStroke, line width=1.2pt},
  flowamber/.style={-{Stealth[length=2.2mm]}, draw=amberStroke, line width=1.2pt},
  flowgreen/.style={-{Stealth[length=2.2mm]}, draw=greenStroke, line width=1.0pt},
}

\usepackage[colorlinks=true, linkcolor=blue, citecolor=blue, urlcolor=blue]{hyperref}
\usepackage{times}
\usepackage{epsfig}
\usepackage{graphicx}
\usepackage{float}
\usepackage{wrapfig}

\usepackage{amsmath,amssymb}

\usepackage{bm,xspace}
\usepackage{comment}
\usepackage{verbatim}
\usepackage{multirow}
\usepackage{multicol}
\usepackage{balance}
\usepackage{url}
\usepackage{booktabs}
\usepackage{etoolbox,siunitx}
\usepackage{calc}
\usepackage{pifont,hologo}
\usepackage{color}

\usepackage[super]{nth}
\usepackage{nicefrac}
\def\aa{\mathbf{a}}

\def\ee{\mathbf{e}}
\def\ff{\mathbf{f}}

\def\oo{\mathbf{o}}
\def\pp{\mathbf{p}}
\def\qq{\mathbf{q}}

\def\tt{\mathbf{t}}

\def\vv{\mathbf{v}}

\def\RR{\mathbf{R}}

\def\bB{\mathcal{B}}

\def\lL{\mathcal{L}}

\def\bmu{{\bm\mu}}

\def\eps{\varepsilon}

\def\trans{^{\top}}

\DeclareMathSymbol{@}{\mathord}{letters}{"3B}

\newcommand\norm[1]{\left\lVert#1\right\rVert}

\newcommand\timess{\mathbin{\!\times\!}}

\newcommand\mypara[1]{\vspace{3pt}\noindent\textbf{#1.}}

\def\latex/{\LaTeX}
\def\bibtex/{\hologo{BibTeX}}

\usepackage[official]{eurosym}

\usepackage{subcaption}

\usepackage{hyperref}
\hypersetup{
    colorlinks=true,
    linkcolor=magenta,
    filecolor=magenta,
    urlcolor=blue,
    pdftitle={HALO},
    pdfpagemode=FullScreen,
    }

\usepackage[capitalize]{cleveref}
\Crefname{table}{Table}{Tables}
\Crefname{section}{Section}{Sections}
\Crefname{figure}{Figure}{Figures}
\Crefname{algorithm}{Algorithm}{Algorithms}
\Crefname{equation}{Equation}{Equations}
\Crefname{appendix}{Appendix}{Appendices}

\newcommand{\cmark}{\textcolor[rgb]{0.2,0.6,0.2}{\checkmark}}%
\newcommand{\xmark}{\textcolor[rgb]{0.7,0.2,0.2}{$\times$}}%
\newcommand{\pmark}{\textcolor[rgb]{0.6,0.6,0.2}{$\sim$}}%

\newcommand{\sysname}{\textsc{HALO}}
\newcommand{\realSR}{91}  %

\definecolor{drop}{HTML}{B00020}
\definecolor{hl}{HTML}{EAF2FD}
\newcommand{\dn}[1]{\textcolor{drop}{$-#1$}}

\renewcommand{\subsubsection}[1]{\par\refstepcounter{subsubsection}\mypara{#1}}

\begin{document}

\bstctlcite{IEEEexample:BSTcontrol} %

\title{Perceptive Humanoid Motion from Simulation}

\author{\authorblockN{Siyuan Hu, Abulikemu Abuduweili, Yue Meng, Shipeng Liu, Quentin Leboutet\\
\\
Chen Chen, Matthias M\"uller, and Vladlen Koltun} \\
\authorblockA{Apple}
}

\twocolumn[{%
  \renewcommand\twocolumn[1][]{#1}%
  \maketitle
  \vspace{-12pt}
  \centering
    \includegraphics[width=\textwidth]{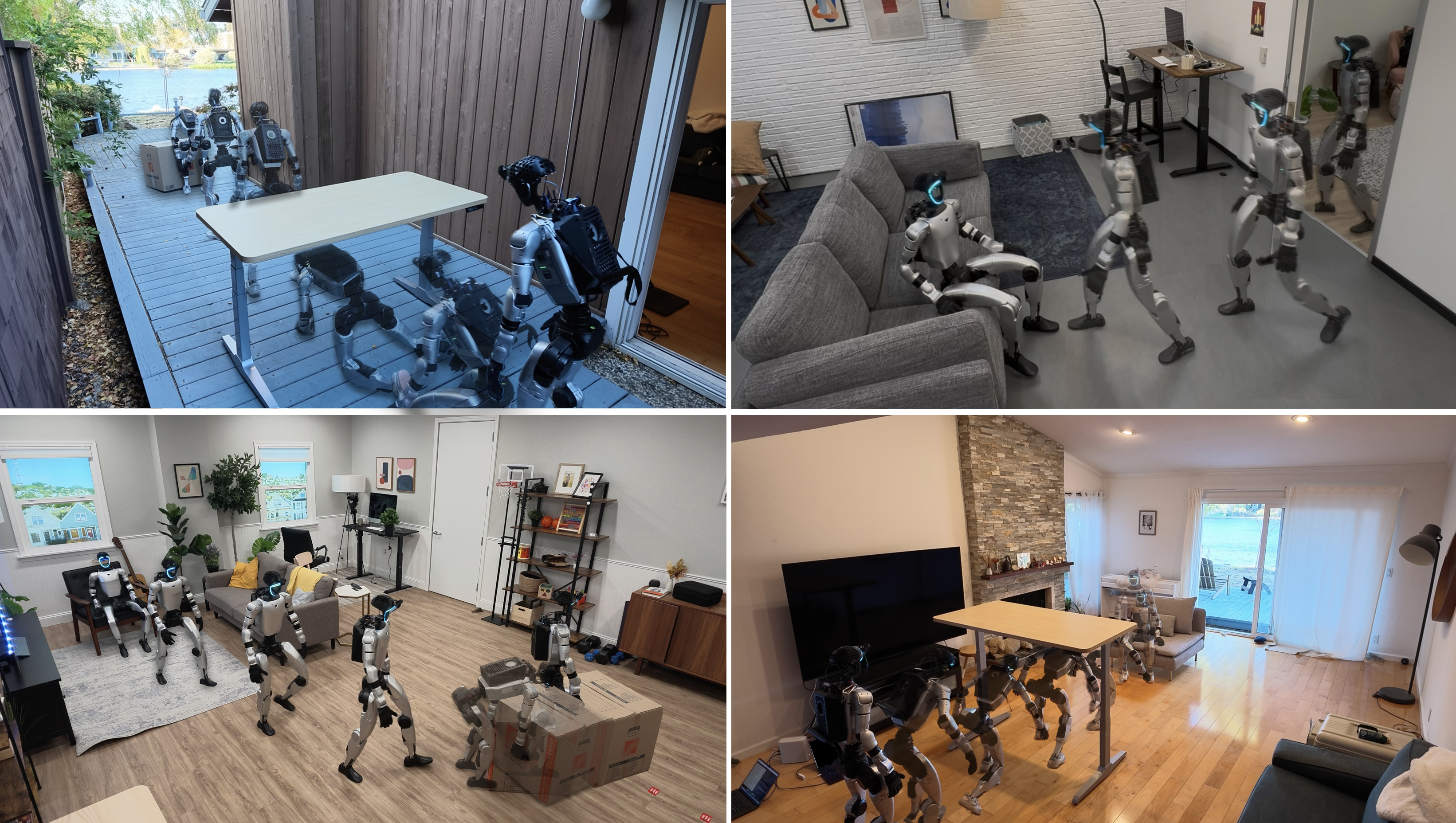}
    \captionof{figure}{\textbf{Autonomous whole-body behavior in real-world environments the robot has
    never seen.} In each environment, the robot is acting in response to a natural-language instruction. The robot sees only its own egocentric camera feed. No scene is scanned and no trajectory is scripted.}
    \label{fig:teaser}
    \vspace{6pt}
}]

\begin{abstract}
Humanoid robots have the potential to automate a wide range of physical tasks. Progress toward generalist humanoid autonomy, however, remains constrained by a data bottleneck: mapping pixels and free-form language to low-level joint commands requires paired supervision, and no corpus of such data exists at sufficiently large scale.
Some approaches rely on human teleoperation and wearable capture devices, which require at least an hour of human time to produce an hour of data.
We present \sysname, a humanoid autonomy system powered by a pipeline that synthesizes vision-language-motion data across a thousand procedurally generated homes. The corpus scales with compute rather than human labor.
\sysname{} acquires a repertoire of physical skills, including sitting on chairs, navigating to named objects, stepping sideways through half-open doors, crouching and crawling beneath tables, and pushing boxes. It orchestrates these skills to solve composite, long-horizon tasks. Without ever seeing a real home during training, \sysname{} transfers zero-shot to the physical world, achieving a \realSR\% success rate across 360 trials in six unmapped real-world environments. See \url{https://apple-aiml-research.github.io/ml-halo/} for code and videos.
\end{abstract}

\section{Introduction}
\label{sec:intro}

\begin{figure*}[!t]
  \centering
  \includegraphics[width=\textwidth]{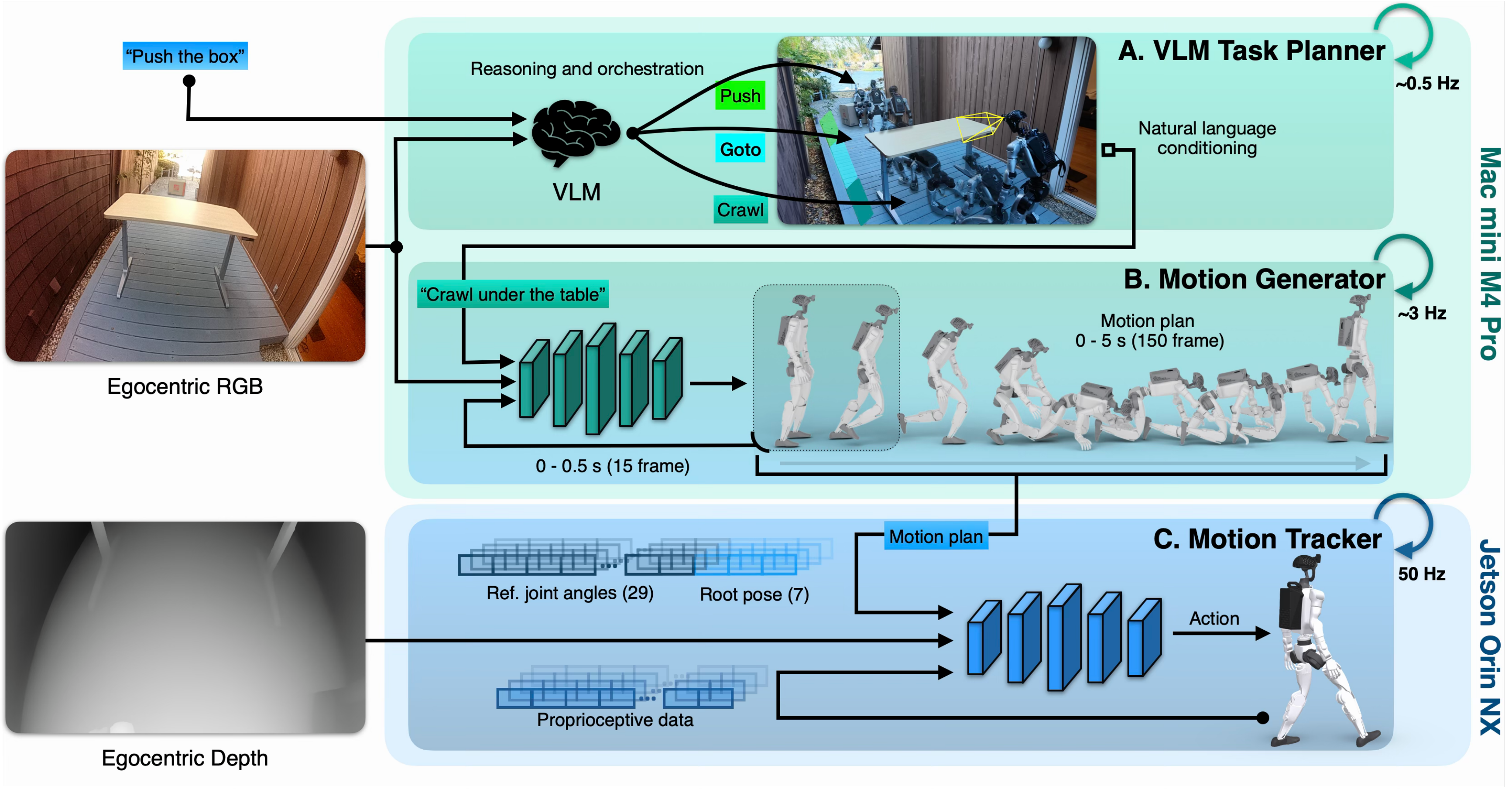}
  \caption{\textbf{Onboard autonomy stack.}
  \textbf{(A)}~A VLM ($\sim$0.5\,Hz) processes a natural-language request and egocentric visual input (top left) and outputs a sequence of subtasks. \textbf{(B)}~The motion generator ($\sim$3\,Hz) maps each subtask and the corresponding egocentric RGB frames into a $5\,$s whole-body motion plan. \textbf{(C)}~The
  motion tracker ($50\,$Hz) executes the motion plan, grounding the trajectory into the physical world using egocentric depth (bottom left) and proprioception.}
  \label{fig:architecture}
  \vspace{-3mm}
\end{figure*}

Humanoid robots can somersault, dance, and run. Yet none of them can be invited into an unfamiliar living room and asked to sit down. They are agile but blind: a backflip is choreographed in advance and executed against geometry the robot never has to perceive. The tasks that would make a humanoid useful are not like this. They take place in rooms nobody has scanned, among furniture nobody has annotated, and the robot has to find the chair before it can sit on it. Perception enters twice, and at two levels of abstraction. It is semantic, deciding what to do: which of the objects around me does the instruction refer to? And it is physical, deciding how to move: the seat has to be met where it actually is, and the robot's weight transferred onto it without losing balance.
Neither level is supervised the way a backflip is. A backflip requires only a reference motion clip and a tracker to follow it. Mapping pixels and free-form language to joint commands, however, demands paired supervision: an egocentric image, an instruction, and the precise whole-body trajectory that satisfies it.

In natural language processing, broad generalization has been unlocked by massive data volume~\cite{brown2020gpt3}. But while text and video are abundant, as a byproduct of people writing and filming, whole-body motion with paired context is not. Every single episode must be deliberately produced. Teleoperation systems such as ALOHA~\cite{zhao2023aloha} and TWIST~\cite{ze2025twist}, or wearable capture devices such as UMI~\cite{chi2024universal}, can generate these demonstrations, but they couple data generation with physical labor. Because a person must manually act out every episode, an hour of data costs at least an hour of human time.

In this paper, we introduce \sysname{} (Humanoid Autonomy, Learned in simulation, running Onboard), a humanoid autonomy system powered by a data generation pipeline that produces egocentric vision-language-motion data entirely in simulation. The resulting policy runs fully onboard the physical robot. Simulation unshackles data generation from human labor. A scene can be generated, a motion synthesized, and the corresponding egocentric video rendered with no human intervention. What would have taken hours (and days, and years) of human time can instead proceed autonomously and in parallel, bounded only by compute.
To create a training episode, a motion diffusion model generates the whole-body humanoid motion, a procedural generator supplies the environment, and the simulator situates the motion in the environment to render the robot's sensory inputs. Our training corpus comprises $300$ hours of episodes across a thousand procedurally generated homes, none of which anyone has had to visit or scan. Data scales with compute rather than human time. Quality is maintained by automated verifiers.

The \sysname{} autonomy stack has three layers, illustrated in \cref{fig:architecture}. An orchestrator (an off-the-shelf vision-language model invoked every two seconds) decomposes a free-form language request into a sequence of subtasks, each identifying one skill and the object it applies to. A motion generator running at 3\,Hz combines a VLM-based encoder with a DiT-based decoder and maps an egocentric RGB observation and the subtask to a whole-body kinematic motion plan: five seconds of joint angles and pelvis poses. Finally, a 50\,Hz motion tracker takes this motion plan in conjunction with onboard depth and proprioception, and drives the robot via joint-level actuator commands. The motion tracker is trained by large-scale reinforcement learning to follow motion plans while respecting balance, contact, and actuator limits. The first two layers settle the semantic half of the problem (which chair?); the tracker settles the physical half (meeting the seat where it actually is).

With this stack, \sysname{} autonomously performs perceptive whole-body skills: sitting on chairs, navigating to named objects, stepping sideways through half-open doors, crouching and crawling beneath tables, and pushing boxes. It also orchestrates them into longer multi-step sequences in response to free-form natural-language requests (\cref{fig:teaser}). %

This paper makes three contributions:
\begin{enumerate}\itemsep2pt
  \item \textbf{Data generation pipeline.} \sysname{} produces egocentric vision-language-motion episodes in simulation, with no teleoperation, no capture suit, and no scan of the deployment environment. Data volume is decoupled from human effort and is bounded only by compute. We train across a thousand procedurally generated home environments.
  \item \textbf{Onboard autonomy stack.} A high-level orchestrator determines the next semantic subtask. A mid-level generator then solves this task from the current egocentric view, emitting a five-second whole-body motion plan. A low-level tracker grounds this plan in the physical world, executing the trajectory using onboard depth and proprioceptive feedback. No prior system pairs a perceptive low-level controller with a whole-body motion plan produced by a generative model running entirely onboard.
  \item \textbf{Whole-body autonomy demonstrated in previously unseen physical environments.} Deployed on the physical robot, \sysname{} achieves a \realSR\% success rate over 360 trials across six real-world home environments that were neither seen during training nor scanned or mapped. The system transfers directly from simulation to novel physical environments, with no preparatory scanning of the target environment.
\end{enumerate}

\section{Related Work}
\label{sec:related}

\subsubsection{Motion supervision} Whole-body behavior is typically obtained from recorded human motion, retargeted to the robot's kinematics and tracked on hardware~\cite{peng2021amp, he2025asap, yang2025omniretarget}. Reference trajectories are generally sourced from physical demonstrations, whether captured via optical rigs, teleoperation suits~\cite{ze2025twist}, handheld capture devices~\cite{chi2024universal}, or raw video~\cite{weng2025hdmi}. \sysname{} bypasses this manual data collection by generating trajectories directly, using a generative model trained on human motion~\cite{rempe2026kimodo}. A generated trajectory may not be physically realizable. It may slide a foot, penetrate the floor, or float off the ground.
Prior methods handle this physical infeasibility at different stages: it can be constrained inside the sampler~\cite{yuan2022physdiff}, caught during a physics-based rollout~\cite{li2026tango}, or repaired post-hoc within the dataset~\cite{xu2025parc}. \sysname{} applies geometric verification immediately after sampling, discarding invalid trajectories (\cref{sec:datagen}).

\subsubsection{Training environments}
Generalizing to unseen homes requires diverse training environments, sourced via reconstruction or procedural generation. If the deployment target is known, reconstruction is highly effective: a brief video scan costs $15\times$ less than teleoperation~\cite{kim2026legs} and often outperforms unrelated diversity~\cite{deitke2023phone2proc}. Consequently, systems like VLK train and evaluate exclusively inside pre-scanned rooms~\cite{wang2026vlk}. True autonomy, however, demands operating in previously unseen, unmapped spaces. \sysname{} leverages procedural generation to obtain diverse training environments~\cite{deitke2022procthor, raistrick2024infinigenindoors}. Prior methods impose strict limits: GRAIL varies terrain but restricts object interactions to authored templates~\cite{xie2026grail}, while FetchMan generates $150{,}000$ scenes solely for a single reach-and-pick task~\cite{rayyan2026fetchman}. \sysname{} removes these restrictions, training across a thousand procedurally generated homes with different architectural layouts and furniture (\cref{tab:comparison}).
\begin{table}[t]
  \centering
  \caption{\textbf{Comparison of teleoperation-free humanoid systems.} Blocks group capability (top), training data (middle), and evaluation (bottom). \emph{Training envs.} counts the environments an episode is staged in; ZeroWBC learns from human egocentric video. \emph{Trials} are summed over tasks. \pmark{} marks a capability present in part, or an evaluation on novel objects but not environments; \emph{n/a} marks a quantity that does not apply; \emph{--} indicates that the quantity was not reported in the paper. \sysname{} is unique in chaining skills online and integrating onboard exteroception directly into the motion tracker.}
  \label{tab:comparison}
  \setlength{\tabcolsep}{2pt}
  \renewcommand{\arraystretch}{1.1}
  \begin{tabular}{@{}lccccc@{}}
    \toprule
     & ZeroWBC & VLK & GRAIL & TANGO & \textbf{\sysname{}} \\
     & \cite{yang2026zerowbc} & \cite{wang2026vlk} & \cite{xie2026grail} & \cite{li2026tango} & \textbf{(ours)} \\
    \midrule
    Free-form language & \cmark & \xmark & \xmark & \cmark & \cmark \\
    Object interaction & \cmark & \cmark & \cmark & \xmark & \cmark \\
    Multi-step tasks & \xmark & \xmark & \xmark & \xmark & \cmark \\
    Perceptive tracker & \xmark & \xmark & \pmark & \xmark & \cmark \\
    \midrule
    Training envs. & n/a & 2 & 2 & 578 & 1{,}000 \\
    \midrule
    Held-out envs. & \pmark & \xmark & \pmark & \cmark & \cmark \\
    Real trials\,/\,envs. & 300\,/\,-- & 120\,/\,2 & 100\,/\,-- & 45\,/\,3 & \textbf{360\,/\,6} \\
    \bottomrule
  \end{tabular}
  \vspace{-5mm}
\end{table}

\subsubsection{Hierarchical whole-body control} Systems mapping language to joint commands can be classified in terms of the interface between semantic reasoning and physical control. End-to-end approaches omit this interface entirely, predicting joint targets directly from language or vision~\cite{xu2024humanvla, he2025viral}, or using generated plans solely to shape the training reward~\cite{kalaria2025dreamcontrol}. Hierarchical systems use explicit interfaces, such as a compact task-space command~\cite{kim2026legs, lin2026hmimicgen}, sparse body keypoints~\cite{yin2025visualmimic}, or a learned latent vocabulary~\cite{xue2025leverb, xie2026grail}. Architectures closest to \sysname{} pass a full whole-body kinematic trajectory~\cite{he2024omnih2o}. Prior methods apply these with planning horizons ranging from one second~\cite{wang2026vlk} to an entire task~\cite{yang2026zerowbc}, and replanning rates ranging from never~\cite{yang2026zerowbc} to 2\,Hz~\cite{li2026tango}. \sysname{} adopts a receding-horizon strategy. It plans five seconds ahead but replans at 3\,Hz, executing only the leading part of a trajectory before a fresh observation supersedes it. Furthermore, \sysname{} expands this hierarchy with a high-level orchestrator that translates a free-form natural language request into a sequence of subtasks, advancing upon success and actively retrying upon failure.

\subsubsection{Perceptive low-level control}   A purely proprioceptive controller is blind to the physical geometry of the environment. Incorporating vision at the low level is a mature line of work~\cite{zhuang2024humanoidparkour, long2024pim, wang2025beamdojo, sferrazza2024humanoidbench, gu2025humanoidsurvey, amadio2026whatmatters}, but these perceptive policies are typically trained for narrow, specialized skills like traversing clutter~\cite{xue2026humanoidpf}. While some frameworks attempt to generalize by selecting~\cite{chen2026llp} or chaining~\cite{lin2026lessmimic} these skills online, they remain constrained by a finite repertoire. Conversely, general-purpose trackers capable of following an arbitrary reference motion lack perception. Current systems that generate dynamic whole-body references pass them to these blind trackers~\cite{yang2025omniretarget, weng2025hdmi, yang2026zerowbc, wang2026vlk, li2026tango}, leaving the robot unable to react if the physical environment deviates from the generated plan. \sysname{} bridges this gap by introducing a perceptive tracker that integrates egocentric depth to follow generated whole-body reference trajectories.

\subsubsection{Sim-to-real evaluation} The true capability of a simulation-trained policy is defined by its evaluation in the physical world: how many distinct real environments it encounters, and whether those environments were seen during training. Prior work reports limited trials in physical environments that were not seen during training (\cref{tab:comparison}). VLK is evaluated exclusively inside two rooms that were previously scanned and used for training the system~\cite{wang2026vlk}. TANGO is tested in three rooms~\cite{li2026tango}, while ZeroWBC and GRAIL report total trial counts without specifying the number of unique environments~\cite{yang2026zerowbc, xie2026grail}. \sysname{} establishes a more rigorous protocol. We evaluate six distinct whole-body skills across six real-world environments over 360 physical trials. Crucially, none of these spaces were scanned, reconstructed, or seen during training, ensuring strict zero-shot transfer to hardware.

\section{Methods}
\label{sec:method}
  \sysname{} maps natural language and egocentric vision to low-level joint commands in previously unseen real-world environments. Deployed entirely onboard, it requires no prior scene reconstruction, external motion capture, or privileged object state. We train on diverse procedurally generated synthetic data, described in \cref{sec:datagen}. The onboard autonomy stack has three layers, shown in \cref{fig:architecture}.
  At the highest layer, a vision-language model (VLM) orchestrates long-horizon behavior by parsing free-form requests into a sequence of linguistic subtasks (\cref{sec:composer}). Bridging semantics and control, a mid-level motion generator maps the active subtask and egocentric RGB into a whole-body kinematic motion plan (\cref{sec:planner}). This plan is a 5-second, $30\,$Hz trajectory specifying $29$ joint angles and a relative pelvis pose per frame ($36$ values total). Finally, a learned motion tracker executes this reference trajectory on hardware using onboard depth and proprioception (\cref{sec:tracker}).
\begin{figure*}[!htbp]
  \centering
  \includegraphics[width=\textwidth]{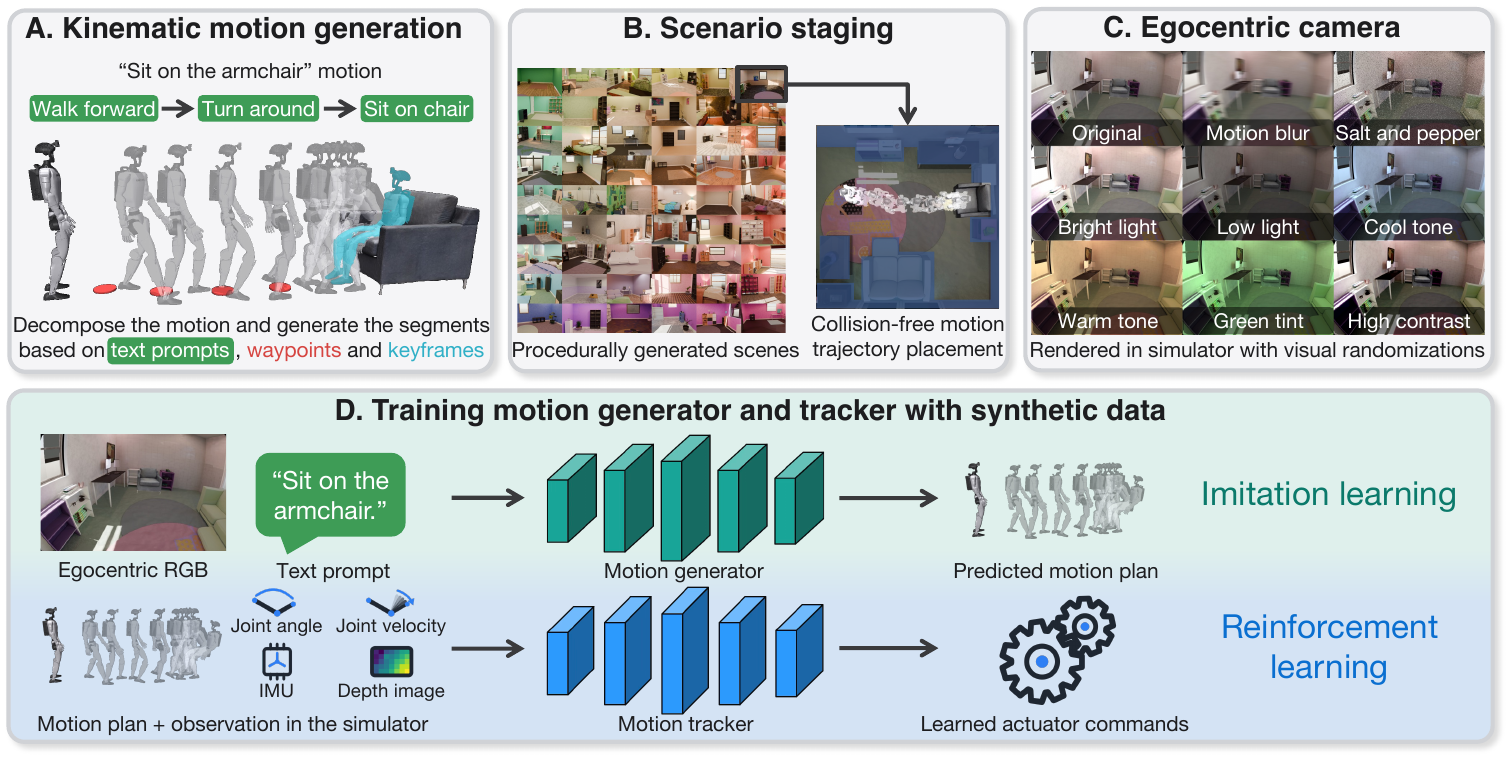}
  \caption{\textbf{Synthetic data generation and training pipeline.} \textbf{(A)}~Each task is decomposed into segments defined by text prompts and keyframe poses. After sampling target asset locations and routing navigation waypoints, a conditional diffusion model synthesizes motion segments from these constraints, stitching them into a continuous whole-body trajectory. \textbf{(B)}~This generated trajectory and its associated assets are embedded into procedurally generated scenes, ensuring collision-free placement. \textbf{(C)}~A simulator replays the kinematic motion to render egocentric vision from the robot's head-mounted cameras. Visual domain randomization is applied to maximize appearance diversity. \textbf{(D)}~The resulting corpus is used for training the autonomy stack, via imitation learning for the motion generator and via reinforcement learning for the motion tracker.}
  \label{fig:data}
  \vspace{-6pt}
\end{figure*}

\subsection{Synthetic data generation}
\label{sec:datagen}

The synthetic data generation pipeline is shown in \cref{fig:data} and detailed in \cref{app:data_generation}.

\subsubsection{Motion generation} We decompose each task into a sequence of motion segments. Each segment is parameterized by a text prompt, optional waypoint guidance, and a terminal keyframe pose defining the object interaction. For example, the task `sit on the armchair' breaks down into `walk forward', `turn around', and `sit on chair'. Waypoints for the `walk forward' segment guide the robot to the front of the seat, while the final segment enforces a seated keyframe anchored to the target asset. This decomposition is written once per task and reused across every procedurally generated environment. See \cref{app:data_generation} for details of the complete segment chain, constraint types, and continuity transformations applied at segment seams. To curate a complete trajectory, we first sample valid locations for the robot and target assets, computing the navigation path via a non-holonomic unicycle planner. We then pass the textual and geometric constraints (waypoints and keyframes) to \textsc{Kimodo}~\cite{rempe2026kimodo}, a kinematic motion diffusion model. \textsc{Kimodo} generates dense whole-body motion for each individual segment, which we then stitch into a single, continuous global trajectory. However, because \textsc{Kimodo} operates purely kinematically, its trajectories may not be physically realizable. Left unchecked, the output can exhibit severe artifacts such as foot sliding, floor penetration, or discontinuous jumps. To improve physical plausibility, %
we generate $N$ candidate trajectories per segment, score them with forward-kinematics geometric verifiers (checking foot skating, floor penetration, flight phases, and joint limits), and retain the highest-scoring candidate.

\subsubsection{Scenario staging} To construct the training corpus, we leverage \textsc{Infinigen}~\cite{raistrick2024infinigenindoors} to procedurally generate $1{,}000$ distinct rooms of three types (living rooms, offices, and bedrooms) with diverse sizes and furniture.
Unlike reconstruction-based pipelines~\cite{wang2026vlk}, this approach requires no physical 3D scans and no manual asset annotation. Environmental diversity scales with compute rather than human labor.
To situate a task within a scene, we first extract a 2D occupancy grid from the generated layout. We then jointly embed the synthesized trajectory and its target asset into the scene by sampling $\mathrm{SE}(2)$ transform candidates, accepting the first configuration that avoids collisions between the kinematic trajectory, the asset, and the surrounding furniture. Because every episode is deterministically defined by a random seed and a lightweight placement record, motion generation and rendering are decoupled: the corpus can be re-rendered at new resolutions, with different camera intrinsics, or inside entirely different physics simulators without regenerating the underlying motion.

\subsubsection{Egocentric rendering} We kinematically replay staged trajectories in scenes generated by \textsc{Infinigen}, rendering egocentric vision through a hardware-matched simulated fisheye lens. To narrow the visual sim-to-real gap, we apply extensive domain randomization encompassing varied color temperature, exposure, contrast, motion blur, and sensor noise (\cref{fig:data}C). Because native path-traced rendering is computationally prohibitive at this scale, we bake global illumination and shading into per-object emissive textures. This one-time preprocessing enables interactive rendering via any standard rasterizer, making the pipeline highly scalable and simulator-agnostic. The resulting corpus comprises $300$ hours of paired video, language, and motion across $1{,}000$ environments. (For later controlled experiments in simulation, episodes are drawn exclusively from completely held-out environments, rather than novel trajectories within training environments.)

\subsection{Motion generator training}
\label{sec:planner}
The motion generator is trained on synthetic data (\cref{sec:datagen}). It takes an egocentric RGB image $I_t$ and a language command $\ell$ as inputs, and produces a full kinematic plan $\hat q_{t+1:t+H}\!\in\!\mathbb{R}^{H\times36}$ over a horizon of $H\!=\!150$ frames (5 seconds at $30\,$Hz; see \cref{app:motion_generator} for architecture and training details).
Each frame $\hat q_\tau\!=\![\hat\theta_\tau; \Delta\hat p_\tau;
\Delta\hat r_\tau]$ stacks the $29$ joint angles with a 3D pelvis translation
and a 4D pelvis quaternion; both are expressed relative to the pelvis frame at $t$, so the output is invariant to global placement.

\subsubsection{Model architecture} The motion generator is an asymmetric encoder-decoder. The encoder is a Qwen3.5-2B VLM that maps an egocentric RGB image and a natural-language command into a joint feature embedding. %
Pretrained VLMs already possess rich, web-scale alignment between visual observations and semantic concepts such as `sit on the chair' or `push the box'. Keeping the encoder frozen leverages these robust representations while requiring only a single forward pass and no optimizer state for its two billion parameters. The decoder is a diffusion transformer~\cite{Peebles2022DiT} that is trained from scratch ($8$ blocks, width $512$, $8$ heads) . It takes as input the VLM feature embedding and generates the dense whole-body motion plan. The decoder is trained via flow matching~\cite{lipman2022flowmatching} and sampled using $15$ Euler steps during inference.

\subsubsection{Asynchronous generation} Closed-loop deployment requires the motion generator to run continuously as new observations arrive. A naive synchronous schedule executes the current plan to completion before invoking the generator on a fresh observation. Because neural inference is not instantaneous, the robot is left without a control signal during the forward pass and must hold its last commanded pose. This stall is inherently unsmooth, fracturing behavior into disjointed segments. It is also physically unsafe: a dynamic system cannot simply pause mid-stride or mid-jump. We therefore run the generator asynchronously. The system computes the next motion plan while the tracker executes the current one. While this eliminates the stall, it introduces a consistency requirement. The current plan spans $[T, T+H)$; the subsequent plan becomes ready at $T+d$, where $0<d<H$. To prevent kinematic jitter at the swap, this newly generated trajectory must smoothly extend the exact state the robot will reach at $T+d$. Methods like real-time chunking (RTC)~\cite{black2025rtc} attempt to close this seam at inference time by using overlapping frames as an inpainting target. This approach suffers from two critical drawbacks. First, continuity is
not mathematically guaranteed. Second, computing this gradient incurs a heavy latency penalty at every denoising step. Instead, we adopt a training-time conditioning formulation~\cite{black2025trtc}. The $P{=}15$ frames ($0.5\,$s) the robot is already committed to executing are passed directly into the generator as clean, noise-free tokens. These occupy the first $P$ frames of the sequence, and the model only denoises the remaining $H-P = 135$
frames ($4.5\,$s). Trained to strictly complete a given prefix, the model internalizes continuity natively. Smooth kinematic transitions become an inherent property of the network rather than a costly inference-time patch, allowing fresh motion plans to stream continuously to the low-level tracker.

\subsection{Motion tracker training}
\label{sec:tracker}
  We ground kinematic plans into physical execution
  in three phases: (1) training privileged single-skill experts, (2) distilling them into a unified depth-based student, and (3) fine-tuning via reinforcement learning with actuator modeling and domain randomization for robust sim-to-real transfer (\cref{app:motion_tracker}).

\subsubsection{Phase 1: Privileged experts}
We begin by training a dedicated PPO expert for each distinct skill. The experts are trained on our synthesized trajectories (\cref{sec:datagen}) and LAFAN1 human motion clips~\cite{harvey2020robust} retargeted to the Unitree G1~\cite{araujo2025gmr}. These experts leverage privileged simulation data, observing proprioception, the next 10 planned frames, and ground-truth scene geometry via local height maps. Each expert outputs joint position targets at $50\,$Hz, which are executed by a low-level PD controller. All experts share an imitation reward that penalizes tracking error on key kinematic links. However, pose imitation alone fails for manipulation. Executing a perfect posture while leaving the target object stationary is a critical failure that purely kinematic rewards cannot detect. Object-centric experts therefore add supplementary terms that reward hand proximity and correct object displacement.

\subsubsection{Phase 2: Multi-teacher distillation} We distill the experts into a single student conditioned only on onboard inputs: the motion plan, egocentric depth, proprioception, and the IMU. During distillation, each simulation instance is supervised by the corresponding expert, so the student never averages conflicting actions from different experts. Training follows a hybrid behavior cloning and Dataset Aggregation (DAgger) strategy \cite{ross2011reduction}, in which trajectories are collected through a mixture of teacher-driven and student-driven rollouts. A 1D convolution encodes the proprioceptive history, a convolutional network processes the depth image, and a residual MLP maps the resulting latents to target joint positions. The student does not receive object or terrain coordinates and must infer that geometry from depth.

\subsubsection{Phase 3: Multi-critic fine-tuning}
Distillation is bounded by teacher competence, and partial observability introduces a residual tracking gap. We therefore apply reinforcement learning directly on the student with depth in the loop. To resolve reward-scale disparity across skills, we use multi-critic PPO: the actor is shared, but each skill maintains an independent critic initialized from its Phase~1 expert. Each critic trains exclusively on samples from its skill and normalizes its advantages independently, preventing high-magnitude returns from dominating the policy gradient.

\subsubsection{Sim-to-real transfer} Beyond domain randomization, we explicitly model actuator physics to minimize the sim-to-real gap. Following OmniXtreme~\cite{wang2026omnixtreme}, we analytically derive joint PD gains from motor rotor inertia to guarantee uniform natural frequencies and damping ratios across all joints, and scale actions by joint torque limits. A random action delay simulates hardware command latency. To protect motor drivers during deep crouches, the reward function directly penalizes excessive negative mechanical power at the knees. This keeps the policy from relying on torques that the simulator silently clips but the real motors cannot deliver.
Remaining sim-to-real gaps are handled by domain randomization, which is kept narrow in Phases 1 and 2 to preserve teacher fidelity, then widened in Phase~3. We randomize physical parameters including surface friction, restitution, base center-of-mass, actuator gains, joint offsets, external impulse pushes, and object mass. Concurrently, the egocentric depth stream is corrupted with simulated dropout, speckle noise, frame latency, and camera mount misalignments. Unlike blind trackers that trip over unmodeled terrain, our depth-conditioned policy leverages this robust training to actively negotiate geometric variation that the upstream kinematic plan failed to anticipate (\cref{tab:ablation}).

\subsection{Orchestration for long-horizon tasks}
\label{sec:composer}
While the generator and tracker excel at atomic, short-horizon behaviors, true autonomy requires satisfying open-ended, long-horizon requests. To achieve this, we top the stack with a VLM-based task orchestrator (\cref{fig:architecture}). Unconstrained natural language presents two fundamental challenges. First, it is inherently underspecified. A command like `sit down' names no explicit target, forcing the system to infer the intended object directly from the visual context. Second, implicit geometric preconditions often expand a single request into a complex sequence. For instance, `sit on the chair' translates to a single subtask if the path is clear, but requires a chained sequence of crouching, approaching, and sitting if a table obstructs the robot. To navigate this, the orchestrator processes the egocentric video stream to resolve ambiguous instructions into an ordered plan of actionable subtasks. It drives this sequence in a closed loop. By continuously monitoring the environment, the VLM evaluates the success of the current action, advances the plan when appropriate, and dynamically retries if a failure occurs. At each step, it passes a single, explicit subtask down to the motion generator for execution (\Cref{app:orchestrator}). By default, the orchestrator runs onboard on the Mac mini using Qwen3.6-35B-A3B (4-bit, MLX), sharing a serving engine with the motion generator (\Cref{app:orchestrator}), to enable operation without internet connectivity. Optionally, it can route requests to frontier cloud models via their APIs for stronger reasoning and planning capabilities.

\section{Experiments}
\label{sec:exp}

\subsection{Data and training}
\label{sec:impl}

The primary training corpus comprises $300$ hours of episodes over $1{,}000$ procedurally generated homes (\cref{sec:datagen}). Synthesizing this data requires $48$~hours across $250$ H100 GPUs, operating without human intervention. The generator trains on $64$ B200 GPUs, while the tracker uses $4$ H100 GPUs. Held-out episodes (for controlled experiments in simulation) are always staged in completely held-out environments, rather than merely novel trajectories within familiar spaces. Consequently, every performance metric represents generalization to an unseen space.

\begin{table}[t]
  \centering
  \small
  \setlength{\tabcolsep}{7pt}
  \renewcommand{\arraystretch}{1.0}
  \caption{\textbf{Task success in held-out environments.} Single-task evaluation is conducted across 100 simulated and 6 real-world environments, while multi-task evaluation uses 50 simulated and 1 real-world environment. Each task is run for 10 trials per environment. Single-task success requires the robot to complete the task without falling or getting stuck. Multi-task success requires completing all subtasks.}
  \label{tab:homes}
  \begin{tabular}{@{}ll cc@{}}
    \toprule
    & & \multicolumn{2}{c}{Success rate (\%)} \\
    \cmidrule(l){3-4}
    Category & Task & Sim. & Real \\
    \midrule
    \multirow{7}{*}{\emph{Single-Task}}
      & \textsc{Go-To}    & 98 & 92  \\
      & \textsc{Sit}      & 84 & 85  \\
      & \textsc{Door}     & 91 & 93  \\
      & \textsc{Crouch}   & 95 & 100 \\
      & \textsc{Crawl}    & 84 & 90  \\
      & \textsc{Push}     & 93 & 88  \\
    \cmidrule(l){2-4}
      & \textbf{Mean}     & \textbf{91} & \textbf{\realSR} \\
    \midrule
    \multirow{4}{*}{\emph{Multi-Task}}
      & \textsc{Door\textrightarrow Go-To}          & 71 & 100 \\
      & \textsc{Crouch\textrightarrow Sit}          & 73 & 60  \\
      & \textsc{Crouch\textrightarrow Go-To\textrightarrow Push}  & 64 & 50  \\
    \cmidrule(l){2-4}
      & \textbf{Mean}                 & \textbf{69} & \textbf{70} \\
    \bottomrule
  \end{tabular}
  \vspace{-10pt}
\end{table}
\subsection{System}
\label{sec:hardware_system}

Our hardware platform is a Unitree G1 with two physical modifications: a custom head mount carrying an OAK-1W camera, and a rear enclosure housing a Mac mini M4 Pro (64\,GB). Perception and control are distributed across these two machines in a fully untethered, asynchronous cascade linked via ZeroMQ (\cref{app:system}). In the outer loop, the OAK-1W streams a $120^\circ$ RGB view ($1280{\times}800$, 30\,fps) to the Mac mini, where the orchestrator and motion generator compute the kinematic reference trajectory. This trajectory serves as the setpoint for the inner control loop running on the robot's factory Jetson Orin NX. The tracker consumes the plan alongside a $55^\circ$ downward-facing Intel RealSense D435i stream ($424{\times}240$, 30\,fps) to issue low-level joint commands. To guarantee strict cycle timing, the tracker is engineered as a C++ real-time process on a \texttt{PREEMPT\_RT} kernel. %
It communicates with the hardware interface via lock-free triple buffers, ensuring the control loop never blocks.
We bypass the D435i's native noisy depth processing, and route the raw stereo IR pair directly into a TensorRT-accelerated stereo matching model~\cite{jing2026lite} on the Jetson Orin NX. The model estimates clean depth at 30\,Hz without the disocclusion or speckle artifacts typical of the native output.

\subsection{Single-task evaluation}
\label{sec:main}
\label{sec:baselines}

These experiments measure whether \sysname{} can execute each task in
environments that never appeared in training, as well as on physical hardware.
We evaluate six tasks:
\textsc{Go-To} (walk to a named object),
\textsc{Sit} (sit on a chair),
\textsc{Door} (step sideways through a half-open door),
\textsc{Crouch} and \textsc{Crawl} (pass under a low obstacle),
and \textsc{Push} (push a box).
Each task is evaluated in $100$ simulated and $6$
real-world environments, with $10$ trials per task per environment~-- $6{,}000$ simulated and $360$ real-world trials in total.
\cref{tab:homes} reports the mean success rate for each task.
The policy achieves high single-task transfer (91\% sim, \realSR\% real).
Some of the real-world environments are shown in \cref{fig:teaser}. \Cref{app:experimental_details} provides more details on the real-world experiments.

\subsection{Multi-task evaluation}
\label{sec:orchestration_results}

To evaluate long-horizon competence, we run the full system with the orchestrator across three representative multi-task sequences in both simulation and the real world.
The results are summarized in \cref{tab:homes}.
Overall, the VLM's task decomposition is robust for the two- and three-subtask chains we evaluate. Chained success falls below the product of the single-task rates ($71\%$ against $89\%$ for \textsc{Door\textrightarrow Go-To}), so the subtasks are not independent.

Finer-grained analysis indicates that the pose in which one subtask ends is an additional factor in multi-task success. Although domain randomization during data generation covers different initial poses, the base position and heading can sometimes be out of distribution. For example, in the real-world \textsc{Crouch\textrightarrow Go-To\textrightarrow Push} test, 3 of 5 failed trials (60\%) end the crouch with the head turned away from the box; the robot never brings it into view and is distracted by other objects. Similarly, two trials of \textsc{Crouch\textrightarrow Sit} end up trying to sit on the sofa on the right instead of the chair because of the ultimate pose of the crouching segment. The same effect appears in simulation for \textsc{Door\textrightarrow Go-To}: walking through the door succeeds 85\% of the time within the chain, but because the door pose varies across scenes, the exit heading varies with it, and the target object can leave the field of view entirely, yielding an 83\% conditional success rate for \textsc{Go-To} and a 71\% success rate overall.
These cases share a single cause (the next subtask begins with its target unobserved), which suggests adding basic recovery behaviors such as search and turn-in-place to make chaining robust.

\begin{table}[t]
    \centering
    \small
    \setlength{\tabcolsep}{6.2pt}
    \renewcommand{\arraystretch}{1.0}
    \caption{\textbf{Stage-wise ablations.} \textsc{Sit} success over $100$ held-out simulated environments. $10$ trials each. Each block varies one stage; each bold row is \sysname{}'s setting (full system), and $\Delta$ is the change in success rate relative to it.}
    \label{tab:ablation}
    \begin{tabular}{@{}ll cc@{}}
      \toprule
      Stage & Configuration & Success rate (\%) & $\Delta$\\
      \midrule
      \multirow{4}{*}{\emph{Training scenes}}
        & \textbf{1{,}000} & \textbf{84} & \\
        & 100 & 81 & \dn{3}\\
        & 10  & 64 & \dn{20}\\
        & 1   & 43 & \dn{41}\\
      \midrule
      \multirow{3}{*}{\emph{Real-time chunking}}
        & \textbf{Training-time} & \textbf{84} & \\
        & Inference-time & 66 & \dn{18}\\
        & w/o & 24 & \dn{60}\\
      \midrule
      \multirow{2}{*}{\emph{Generator decoder}}
        & \textbf{Flow matching} & \textbf{84} & \\
        & Regression & 19 & \dn{65}\\
      \midrule
      \multirow{2}{*}{\emph{Motion tracker}}
        & \textbf{w/ depth} & \textbf{84} & \\
        & w/o depth & 74 & \dn{10}\\
      \bottomrule
    \end{tabular}
    \vspace{-10pt}
  \end{table}
\subsection{Ablation study}
\label{sec:ablation}
Table~\ref{tab:ablation} isolates the contribution of each stage while holding the rest of the system fixed.
First, we vary the number of environments in which the motion is staged from $1$ to $1{,}000$, holding all other stages fixed and training policies jointly on all six tasks. This isolates the effect of environment diversity from motion quantity. Scaling the number of environments from 1 to 100 sharply boosts the success rate ($43\%$ to $81\%$), while scaling further yields 3 additional percentage points.
Next, without real-time chunking (RTC), the motion plan jumps at every swap and the success rate falls to $24\%$. Inference-time RTC smooths the transition through a guidance gradient, but only encourages continuity rather than guaranteeing it; it introduces additional inference latency and still
trails training-time conditioning by $18$ percentage points ($66\%$ against $84\%$).
Replacing the flow-matching DiT with a regression decoder using the same transformer backbone lowers the success rate to $19\%$. RTC does not directly
apply to a non-diffusion model, so we lose continuity in the asynchronous
generation setting. The regression decoder also overfits easily: outside the training
distribution, the plan retains only the coarse root path, losing joint detail,
resulting in foot skating.
Finally, removing depth leaves only proprioception, lowering the success rate to $74\%$.

\section{Conclusion}
\label{sec:conclusion}

We show that simulation can serve as the data engine for humanoid
autonomy. Simulation removes the human operator from the data loop, unshackling data generation from human labor.
The resulting stack transfers zero-shot to real environments that were neither seen during training nor scanned or mapped.

\subsubsection{Limitations and future work}
First, our pipeline removes the operator from data collection, but not from task
design: adding a new physical task still requires a human to write its prompt
and author the keyframes of its motion sequence, so the repertoire grows only as
fast as a person can specify additional tasks.
Second, the robot's sensory input is largely based on vision, which does not provide direct perception of contact: once an object leaves the camera's field of view or is occluded, the policy has no signal that it is still there, and contact-rich tasks degrade accordingly. Tactile sensing would be a natural solution. Third, the
interface between the motion generator and the motion tracker is narrow; the tracker follows the kinematic plan without knowing what the plan is for. Communicating the intended future state of the
world would be a natural next step. Finally, our
orchestration layer is relatively simple;
as the repertoire of tasks grows, we expect to need a more agentic system.

\newif\ifshortbib
\shortbibtrue  %

\bibliographystyle{IEEEtran}
\ifshortbib
  \bibliography{bib/strings,bib/strings_short,bib/main}
\else
  \bibliography{bib/strings,bib/main}
\fi

\clearpage
\appendix
\crefalias{subsection}{appendix}
\crefalias{subsubsection}{appendix}
\subsection{Data generation}
\label{app:data_generation}

The training corpus in \cref{sec:datagen} consists of six datasets (one per skill), with
$162{,}698$ episodes and $296.3\,$h of data at $30\,$Hz in total
(\cref{tab:app-corpus}).  All the skills share the same logic for procedural scene generation (\textsc{Infinigen}), motion generation (\textsc{Kimodo}) and egocentric rendering, whereas each skill has its own sequence of \textsc{Kimodo} calls, quality filters and appearance randomization.
Each episode has one fixed instruction, which is shared by all the episodes of a skill, except for \textsc{Go-To}, which has one instruction per target object. This gives ten instructions in the corpus: ``sit arm\_chair'', ``walk to armchair'', ``walk to basketball'', ``walk to box'', ``walk to man'', ``walk to trashbin'', ``Go through the half-open door'', ``crawl under the table'', ``crouch under the table'' and ``push box''. At run time, the orchestrator converts a free-form request into one of these instructions (\cref{app:orchestrator}).

\begin{table*}[!t]
  \centering
  \caption{\textbf{Training corpus by skill.} Every \textsc{Kimodo} call is guided by a root path. KF denotes full-body keyframe constraints.}
  \label{tab:app-corpus}
  \setlength{\tabcolsep}{3.5pt}
  \renewcommand{\arraystretch}{1.08}
  \scriptsize
  \begin{tabular}{@{}lllllrrc@{}}
    \toprule
    \textbf{Skill} & \textbf{Motion} & \textbf{Keyframes from} & \textbf{Filter} & \textbf{Appearance} & \textbf{Episodes} & \textbf{Hours} & \textbf{Mean length} \\
    \midrule
    \textsc{Sit}    & walk (best of $16$), turn, sit & G1 joint log & $33$ checks, keep best & image randomization & $25{,}349$ & $50.0$ & $7.1\,$s \\
    \textsc{Go-To}  & one walk to an object & standing pose & target in view at start & random dome light & $50{,}000$ & $47.1$ & $3.4\,$s \\
    \textsc{Door}   & one call, $5$ KF & hand-edited poses & none & image randomization & $29{,}207$ & $50.0$ & $6.2\,$s \\
    \textsc{Crawl}  & one call, $6$ KF & phone video, GVHMR, GMR & $21$ checks, best of $20$ & random dome light & $19{,}066$ & $50.0$ & $9.4\,$s \\
    \textsc{Crouch} & one call, $6$ KF & phone video, GVHMR, GMR & $21$ checks, best of $20$ & random dome light & $19{,}369$ & $49.2$ & $9.1\,$s \\
    \textsc{Push}   & walk (best of $10$) $+$ recorded push & motion capture & gait ranking & image randomization & $19{,}707$ & $50.0$ & $9.1\,$s \\
    \midrule
    Total & & & & & $162{,}698$ & $296.3$ & $6.6\,$s \\
    \bottomrule
  \end{tabular}
  \vspace{-6pt}
\end{table*}

\begin{figure*}[!t]
  \centering
  \setlength{\tabcolsep}{1pt}
  \renewcommand{\arraystretch}{0.6}
  \begin{tabular}{@{}ccccc@{}}
    \includegraphics[width=0.196\textwidth]{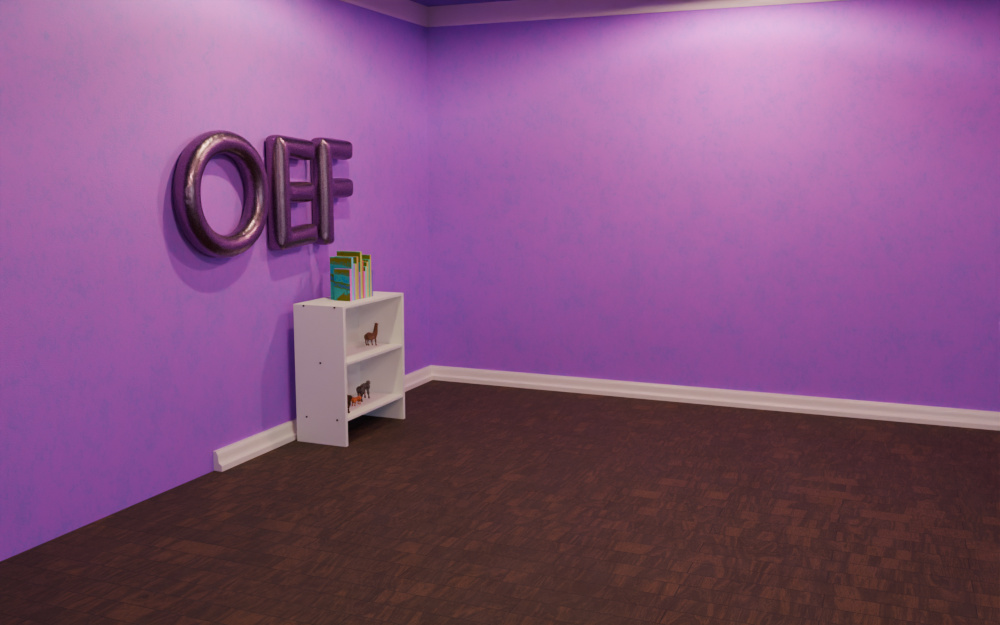} &
    \includegraphics[width=0.196\textwidth]{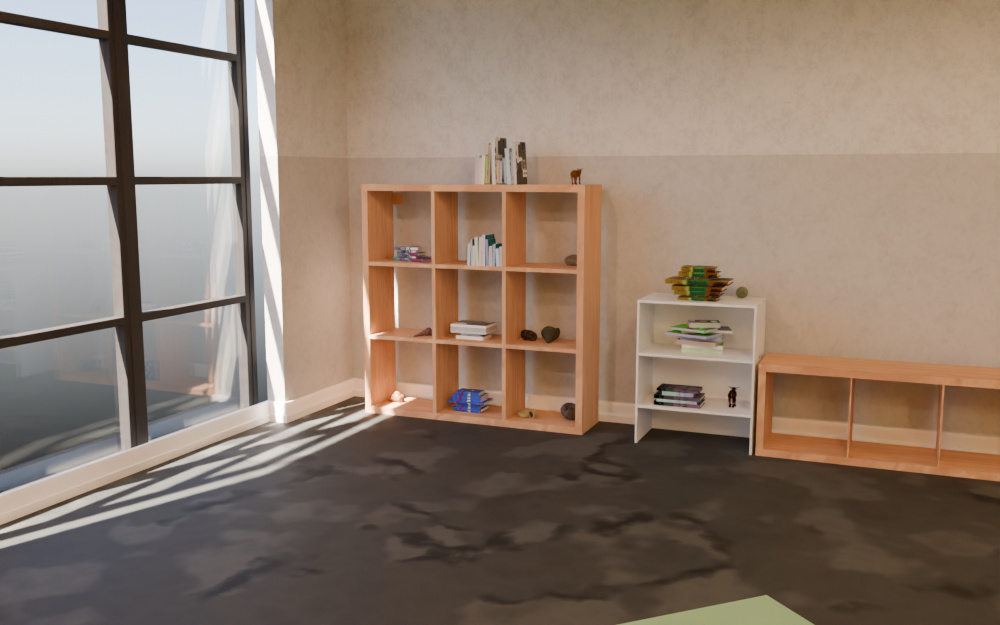} &
    \includegraphics[width=0.196\textwidth]{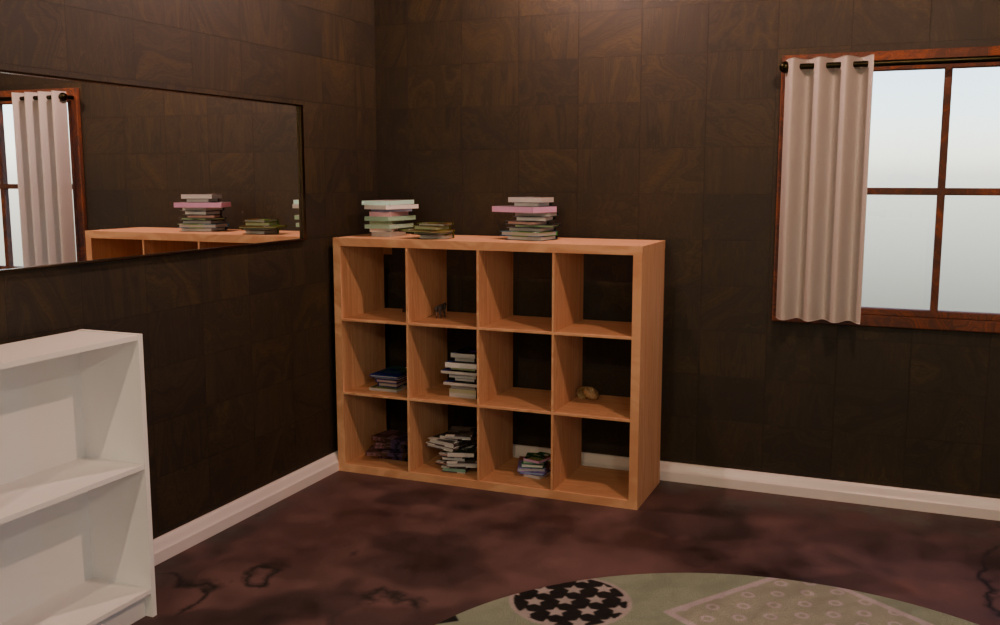} &
    \includegraphics[width=0.196\textwidth]{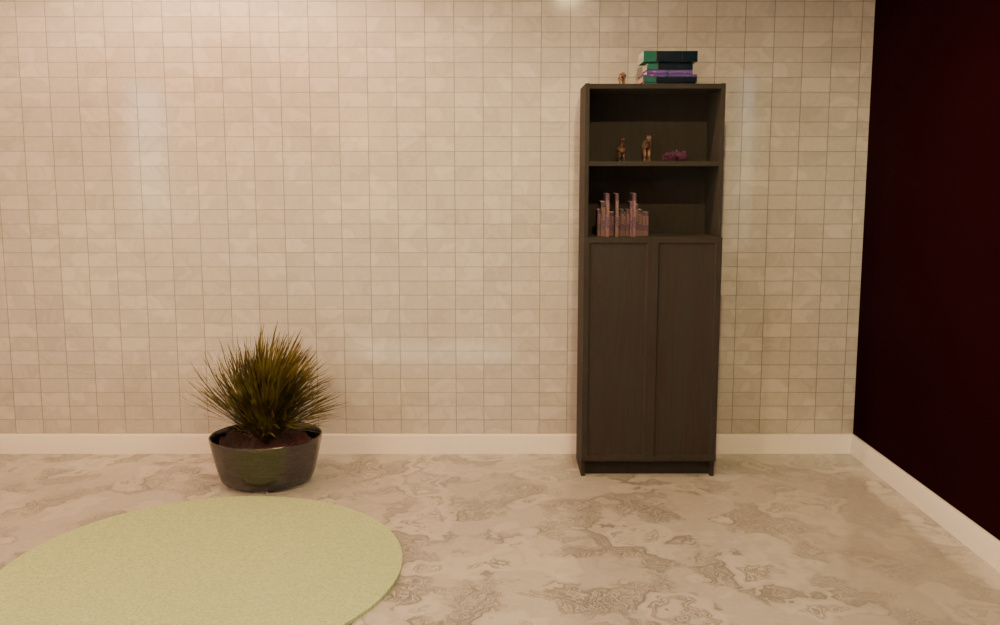} &
    \includegraphics[width=0.196\textwidth]{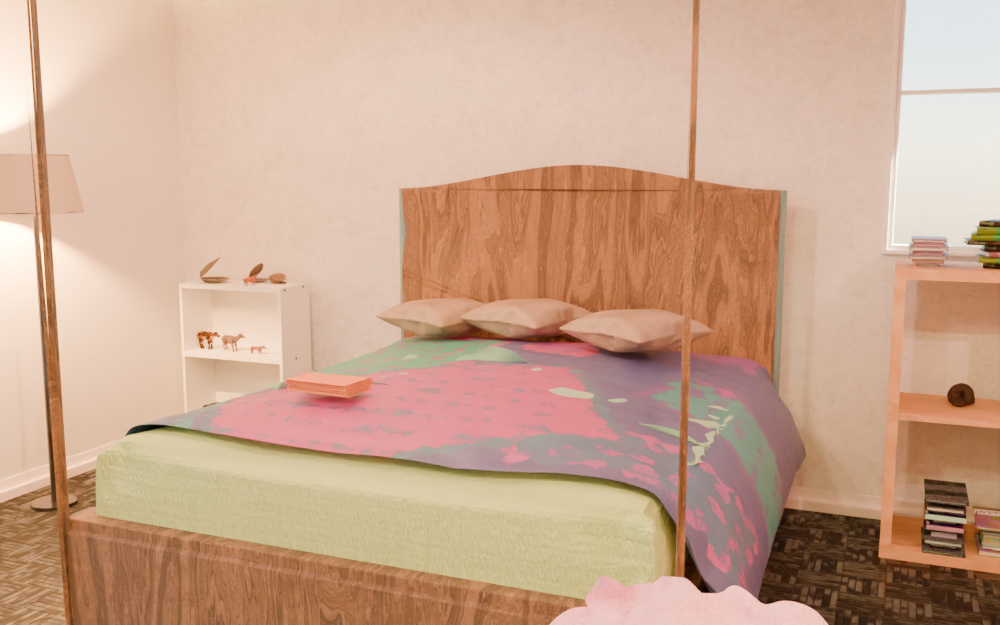} \\
    \scriptsize living room, sparse & \scriptsize living room, dense & \scriptsize office, medium & \scriptsize office, sparse & \scriptsize bedroom, dense \\[3pt]
    \includegraphics[width=0.196\textwidth]{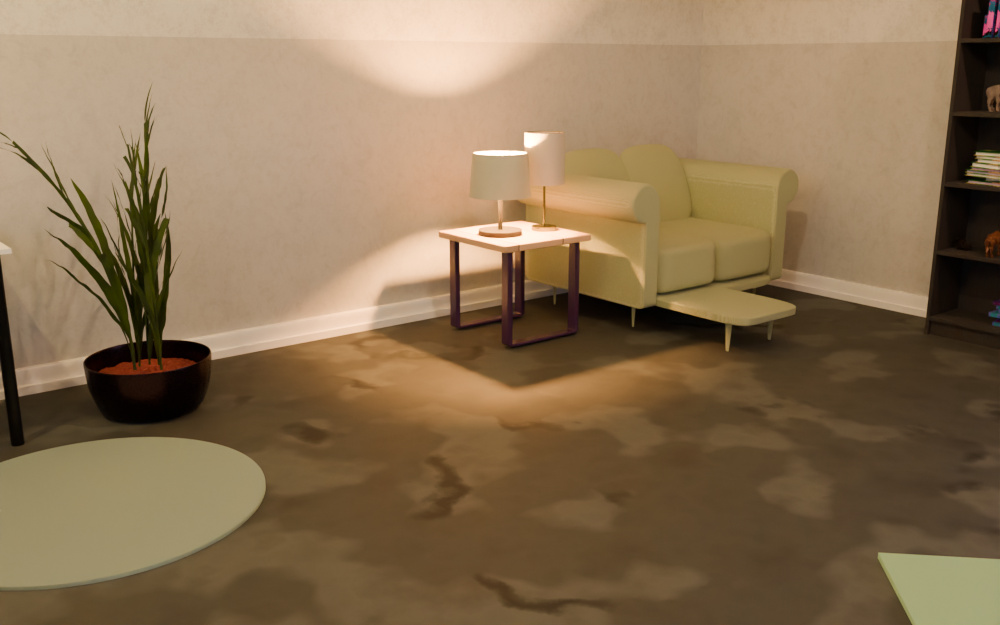} &
    \includegraphics[width=0.196\textwidth]{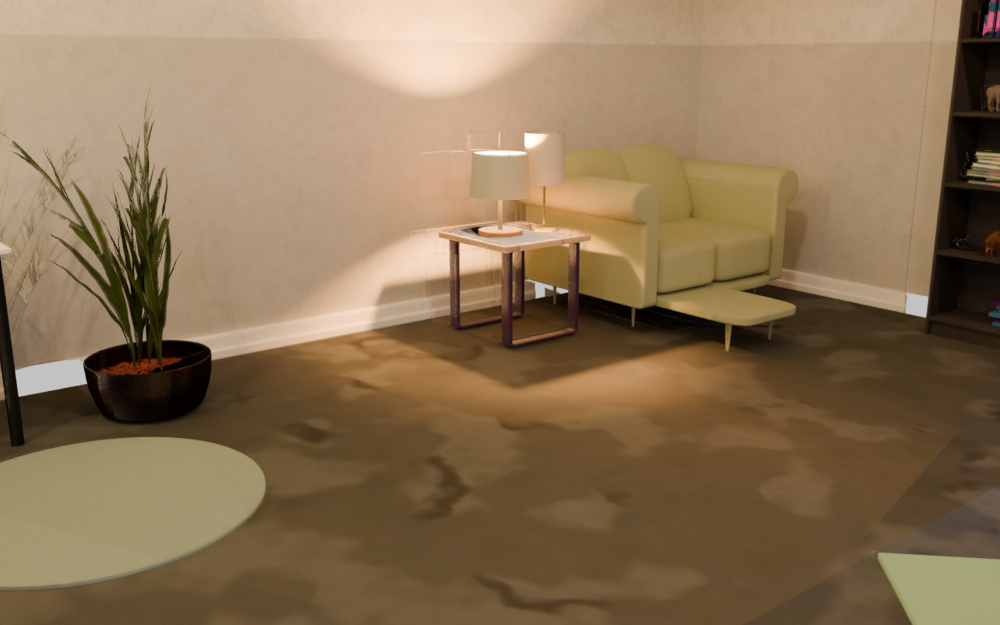} &
    \includegraphics[width=0.196\textwidth]{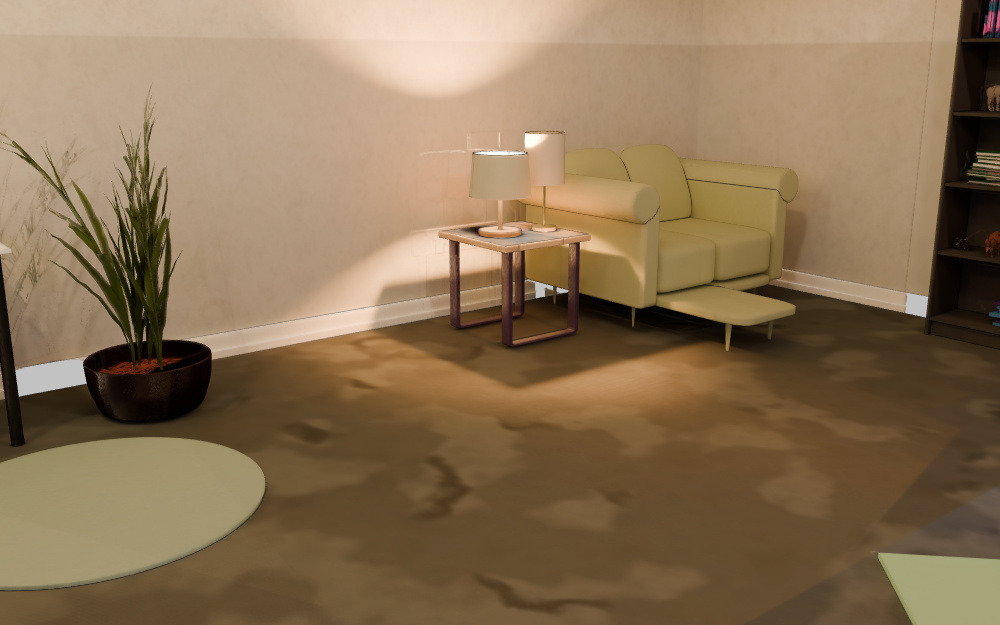} &
    \includegraphics[width=0.196\textwidth]{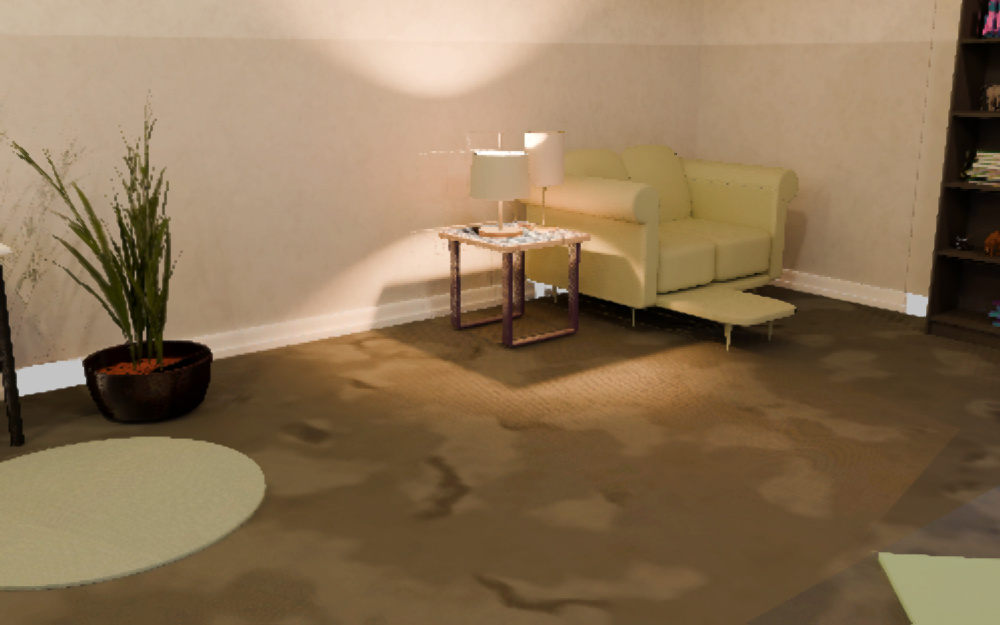} &
    \includegraphics[width=0.196\textwidth]{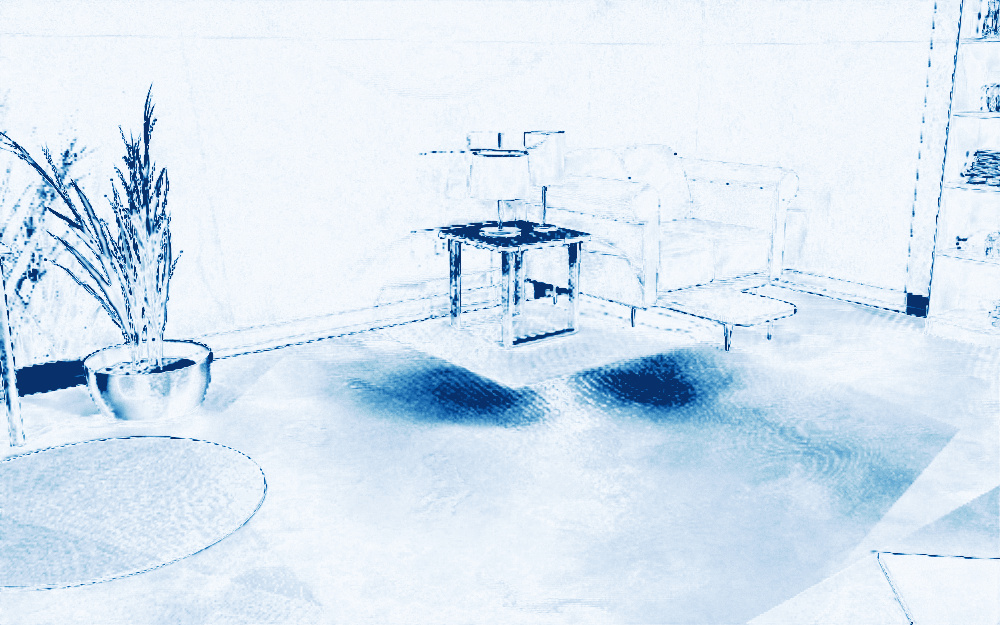} \\
    \scriptsize path traced (reference) & \scriptsize baked, Blender & \scriptsize baked, MuJoCo & \scriptsize baked, Isaac Sim & \scriptsize error, Isaac Sim \\
  \end{tabular}
  \caption{\textbf{Procedural rooms and scene baking.} Top: path-traced training
  rooms from \textsc{Infinigen}. Bottom: a held-out view of a dense living room
  rendered from the bake in three engines, where the mean absolute error relative to path tracing is
  $4.2$ (Blender), $4.5$ (MuJoCo) and $4.7$ (Isaac Sim) grey levels and mostly lies in the light pool of the lamp on the floor and at object edges (error map: white is $0$, dark blue is $32$ grey levels or more).}
  \label{fig:app-rooms}
  \vspace{-8pt}
\end{figure*}

\subsubsection{Scene generation} We use the indoor generator of
\textsc{Infinigen}~\cite{raistrick2024infinigenindoors} to generate single
rectangular rooms, namely living rooms ($45\%$), offices ($30\%$) and bedrooms
($25\%$), with a length of $3.5$ to $9.5\,$m per side. Each room has sparse, medium or dense
furniture ($20$ to $80$ objects), one of four floor materials, up to two rugs, two
or three windows, and plain or decorated walls (\cref{fig:app-rooms}--\emph{top}).
We generate $1{,}200$ rooms, of which $1{,}178$ finish (the other $22$ stop inside
the scene generation process, when its furniture solver hangs or crashes). We use $1{,}000$ rooms
for training and hold out the remaining $178$ rooms, and the simulated evaluation
uses $100$ of the held-out rooms.
\textsc{Door} needs a doorway, so it uses $966$ two-room and multi-room scenes from
the same generator.

\subsubsection{Scene baking} Path tracing all $32$ million training frames in
Blender Cycles is too slow ($13.2\,$s per frame, compared to
$56$--$87\,$ms in Isaac Sim~\cite{gao2026isaacsim} with real-time ray tracing), so we bake
the lighting of each room only once.
We first use Cycles to bake the direct and indirect lighting of every object into
a texture atlas with $128$ samples per pixel. The atlas resolution grows with the object surface area (about $1{,}024$ pixels per meter, up to $4{,}096$ pixels).
Then we project eight path-traced views onto the atlases to correct the texels
they cover, and finally we replace every material with an unlit emissive one.
Since a baked texture cannot represent view-dependent reflections, we make all
the wooden surfaces matte, which halves the bake error on a test room.
The bake takes $38.5\,$minutes per room (median) on a single
GPU.
We measure the bake error as the mean absolute difference between a baked render and
the path-traced render of the same view, in grey levels of an 8-bit image ($0$ to $255$).
For reference, two path-traced renders of the same view with different random samples
already differ by $1.05$ grey levels. Over all rooms, the median error is $2.0$ on the
eight projected views and $3.8$ on $16$ held-out views that the bake never used, i.e.,
on a new view a pixel is off by $1.5\%$ on average (\cref{fig:app-rooms}--\emph{bottom}).
This matters because every training image is rendered from the bake, so the bake error
adds to the visual gap between simulation and the real camera. The bake also carries over
to other renderers: MuJoCo~\cite{todorov2012mujoco} and Isaac Sim render it within $1.5$
and $2.0$ grey levels of Blender, so we can render the corpus in a fast simulator without
baking again. The remaining errors mostly come from glass, mirrors and screens, whose
appearance depends on the view. We leave the ceiling out of the bake, since a closed room
comes out too dark. The head camera is pitched down by $25^\circ$, so the ceiling only
appears near the top edge of the images.

\begin{figure*}[!t]
  \centering
  \setlength{\tabcolsep}{1pt}
  \renewcommand{\arraystretch}{0.6}
  \begin{tabular}{@{}cccccc@{}}
    \includegraphics[width=0.1635\textwidth]{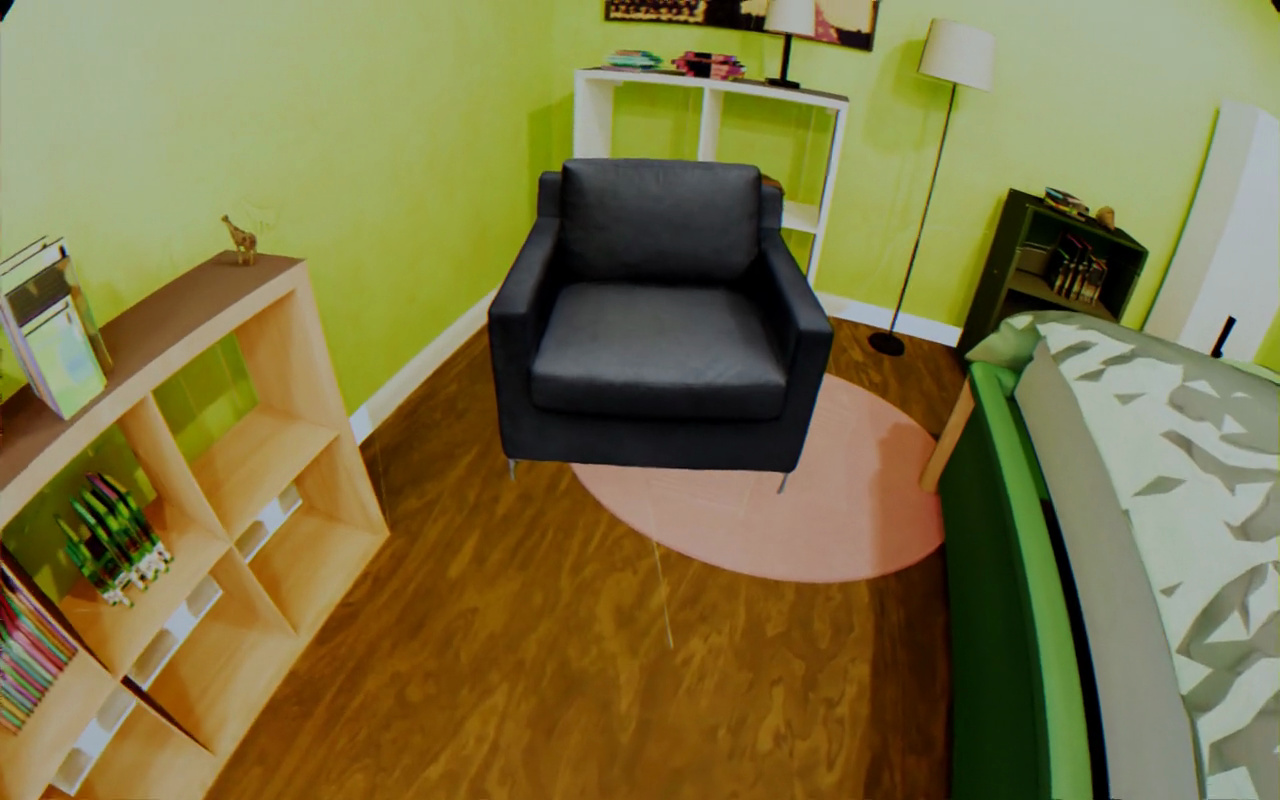} &
    \includegraphics[width=0.1635\textwidth]{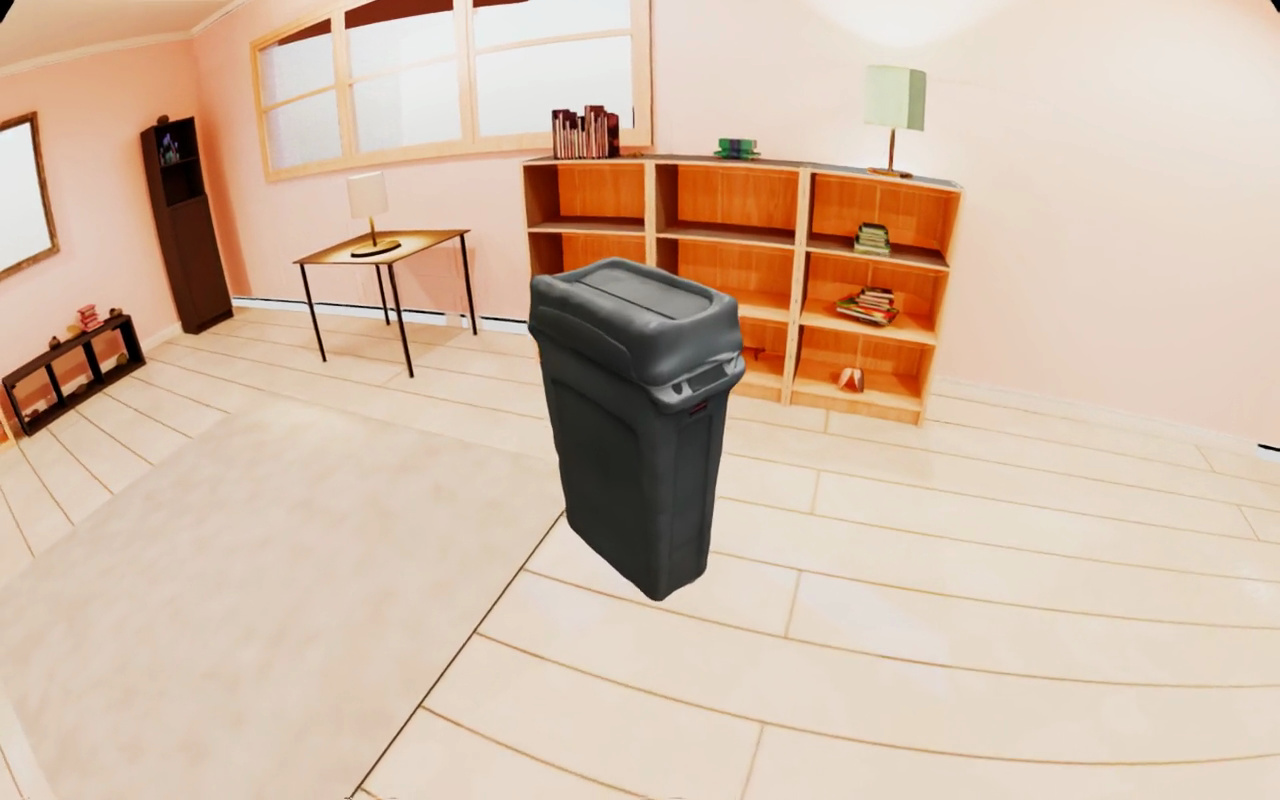} &
    \includegraphics[width=0.1635\textwidth]{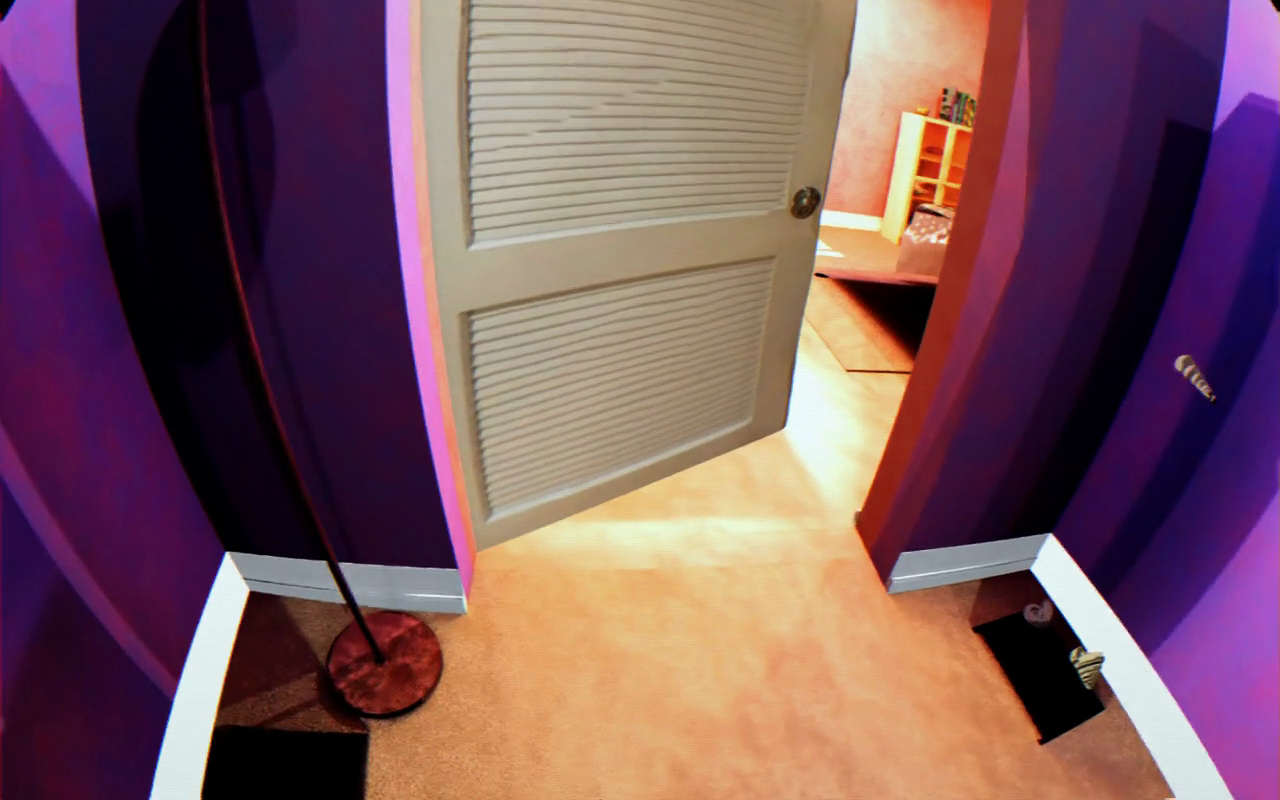} &
    \includegraphics[width=0.1635\textwidth]{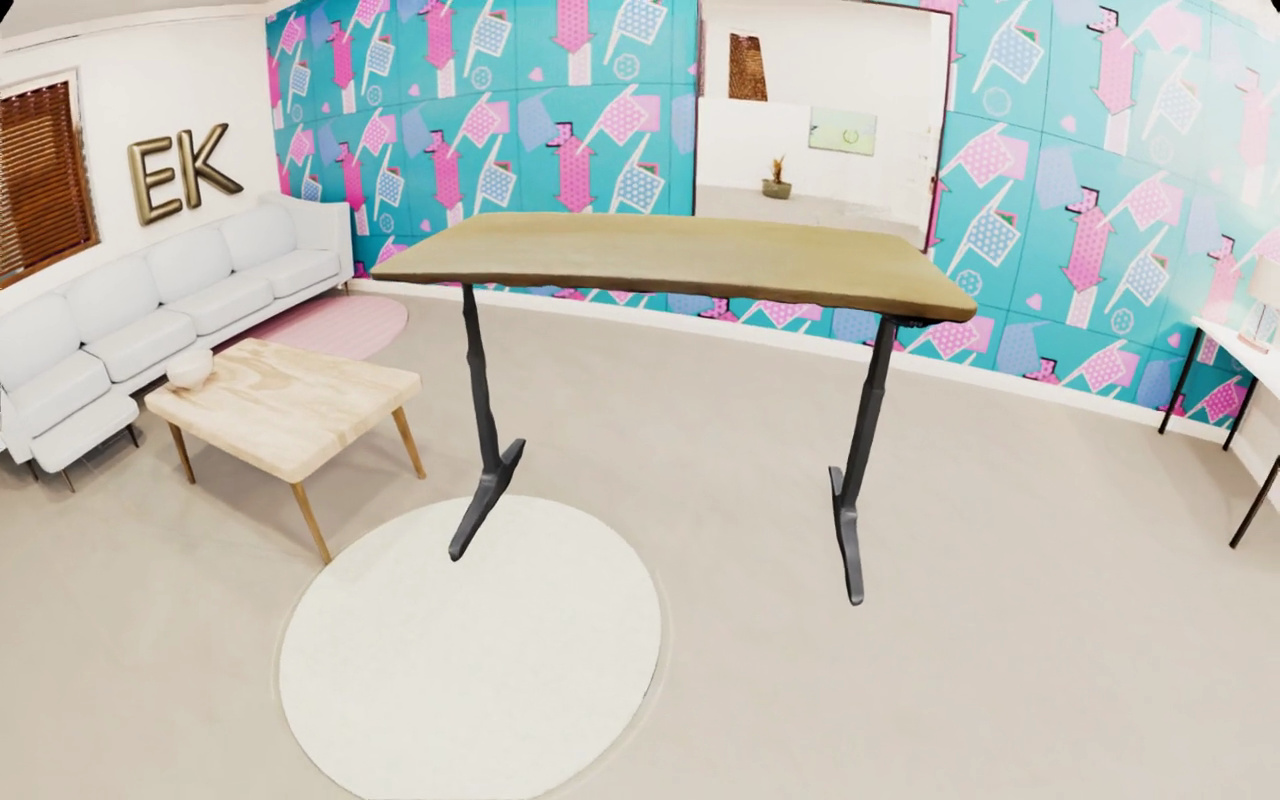} &
    \includegraphics[width=0.1635\textwidth]{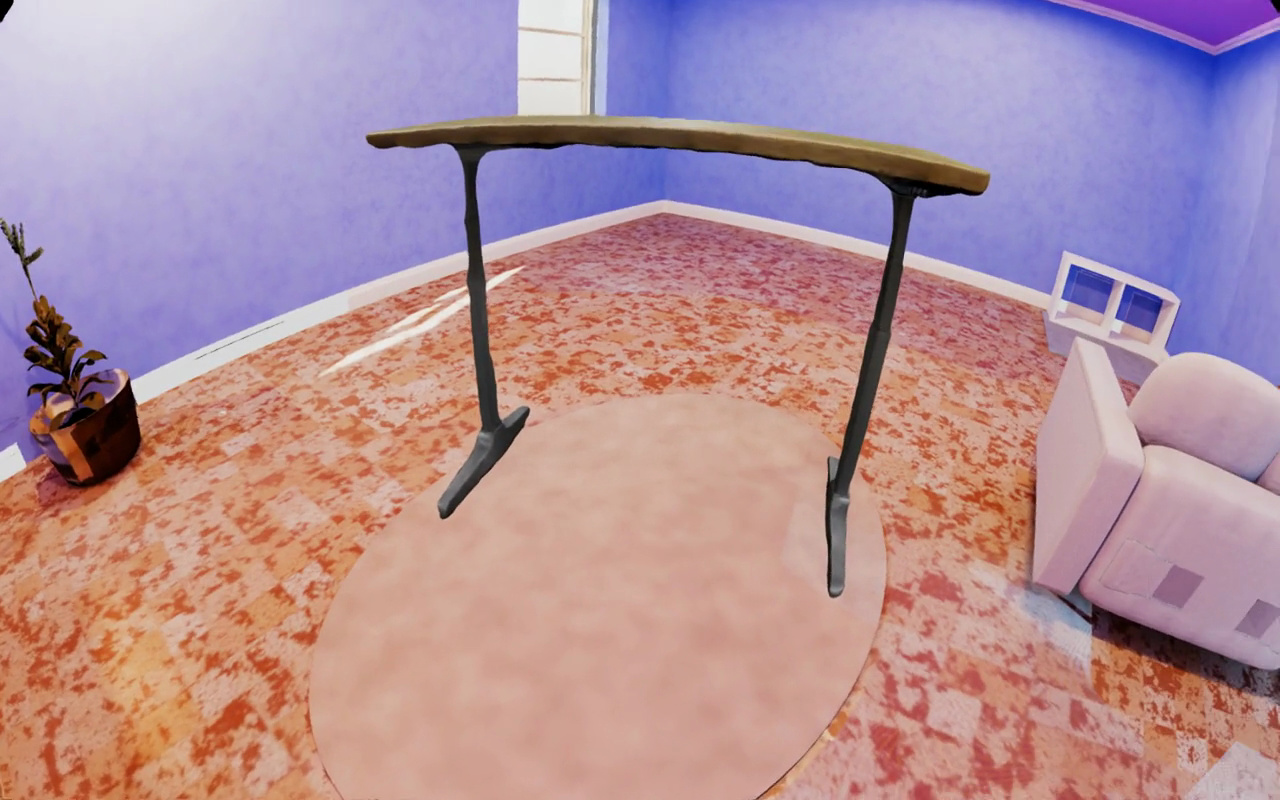} &
    \includegraphics[width=0.1635\textwidth]{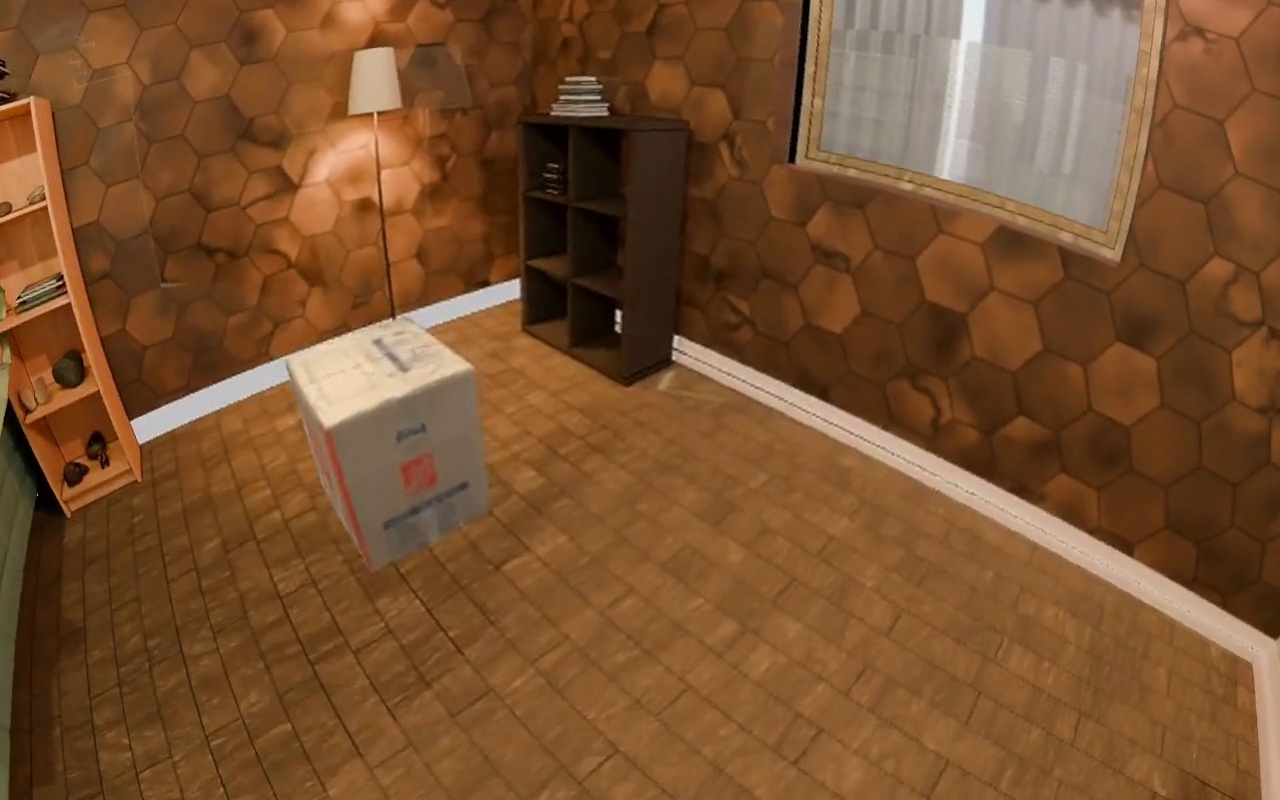} \\
    \scriptsize\textsc{Sit} & \scriptsize\textsc{Go-To} & \scriptsize\textsc{Door} & \scriptsize\textsc{Crawl} & \scriptsize\textsc{Crouch} & \scriptsize\textsc{Push} \\
  \end{tabular}\\[4pt]
  \includegraphics[width=0.86\textwidth]{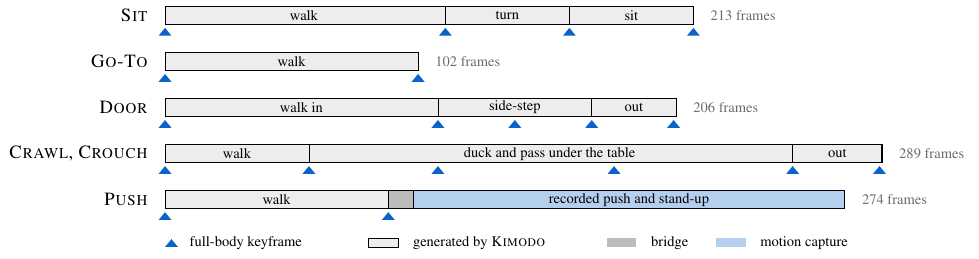}
  \caption{\textbf{Episodes of the six skills.} Top: training frames from the
  egocentric fisheye camera ($1280\timess800$). Bottom: a typical episode of each
  skill on the same time axis (blue triangles are full-body keyframes), where only
  \textsc{Sit} chains several \textsc{Kimodo} calls.}
  \label{fig:app-skills}
  \vspace{-8pt}
\end{figure*}

\subsubsection{Kinematic motion generation} We use \textsc{Kimodo}~\cite{rempe2026kimodo} to generate the
motions, with a classifier-free guidance weight of $2.0$ on both the text and the
constraints (root paths and full-body keyframes in \cref{fig:app-skills}). For
\textsc{Sit}, we generate the whole episode with the chair at the origin before
choosing a room. The robot starts $1.5$ to $3\,$m away from the seat (within
$\pm70^\circ$ in front of the chair), and a unicycle model plans a path to a point
$0.55\,$m in front of the seat. We then call \textsc{Kimodo} three times: a walk
guided by $20$ waypoints, a turn with $50$ frames, and a sit with $50$ frames that
ends at a seated pose from a joint-state log of the real G1. Each segment starts
from the last frame of the previous segment, and we smooth each transition with a
linear cross-fade over three frames on each side. \textsc{Go-To} is a single walk to one of
five objects (trash bin, box, basketball, armchair and a standing person). The walk starts $2$ to $4\,$m away from the object, with the object in view. \textsc{Door} is a
single call where the robot walks to a door opened by $30^\circ$ to $45^\circ$,
side-steps through the gap with three step-together stances, and walks out.
\textsc{Crawl} and \textsc{Crouch} are also single calls, guided by six keyframes
from phone videos of a person going under a table. We lift the videos to 3D with
GVHMR~\cite{shen2024gvhmr}, retarget them to the G1 with GMR~\cite{araujo2025gmr},
and edit the keyframes by hand. \textsc{Crawl} goes on hands and knees under tables $0.70$ to $1.05\,$m high, whereas \textsc{Crouch} goes in a deep stoop under tables $1.05$ to $1.40\,$m high. \textsc{Push} connects a generated walk to a motion-capture recording of a person pushing a $0.46\timess0.50\timess0.56\,$m box, with a
bridge of $9$ to $11$ frames in between.

\subsubsection{Filtering} Since \textsc{Kimodo} is purely kinematic, we score the
candidates on their forward kinematics. For \textsc{Sit}, we use $33$ checks
(e.g., foot skate, hops, flight phases, pauses, torso lean, joint limits and the
final seated height), and a candidate passes if it fails none of the $23$ hard
checks and has a weighted score of at least $0.75$. We first sample $16$ walks in a
batch and keep the best one, then generate the turn and the sit. If the turn or
the sit still fails after two retries, we resample the walk (up to six rounds) and keep the best chain in the end, even if it still fails. Of the $50{,}000$ generated \textsc{Sit} episodes, $77\%$ pass all the hard checks, and the $25{,}349$ episodes in the corpus are
evenly spaced over all the $50{,}000$ episodes. \textsc{Crawl} and \textsc{Crouch}
use $21$ checks with a pass score of $0.65$, keep the best of $20$ candidates, and drop the episode if no candidate passes. For \textsc{Push}, we sample $10$ walks for each start and keep the one with the gait closest to the recorded walk.

\subsubsection{Scene staging} We place each motion into a room with a rigid 2D rotation and
translation, which is applied to both the robot path and the target object. We
search up to $800$ free cells on a $10\,$cm occupancy grid with $24$ headings, and
accept the first transform where the pelvis (checked every three frames) is at
least $0.35\,$m away from all the obstacles and the target object fits in the
room. Since each episode is fully specified by its seed and placement record, we
can re-render the corpus with another camera or renderer without regenerating the
motion. For \textsc{Door}, we instead align the motion to the doorway and remove the furniture on the robot path.

\subsubsection{Egocentric view rendering} We replay the episodes kinematically and render them in Isaac
Sim~5.1 with real-time ray tracing (\textsc{Push} is rendered in MuJoCo). The
egocentric camera matches the calibrated OAK-1W fisheye camera on the robot, with
$1280\timess800$ pixels and a $128^\circ\timess81^\circ$ field of view, pitched down
by $25^\circ$. Since the baked lighting is fixed, we randomize the appearance either after rendering or through the dome light. \textsc{Sit}, \textsc{Door} and \textsc{Push} randomize the exposure, contrast, tint, motion blur (along the camera motion) and sensor noise of the rendered images. \textsc{Go-To}, \textsc{Crawl} and \textsc{Crouch} randomize the intensity and color of the dome light at render time, which changes the brightness and tint of the room because Isaac Sim also lights the baked textures. For the $50{,}000$ \textsc{Sit} episodes,
generating and rendering the episodes and caching their vision-language features takes about $9\,$h on
$250$ H100 GPUs,
and generating the $1{,}178$ rooms takes $26\,$h on up to $300$ GPUs.

\subsection{Motion generator}
\label{app:motion_generator}

\subsubsection{Network architecture}
\label{app:generator_arch}
The generator follows the encoder-decoder design of \cref{sec:planner}: a frozen
Qwen3.5-2B reads the image and the language command, and a DiT
decoder~\cite{Peebles2022DiT} turns its features into a $150$-frame motion plan.
\cref{tab:generator_hparams} provides the full configuration.

We pass the $224\times224$ image and the language command together in a single prompt to Qwen
and use the last-layer hidden states of all $64$ image tokens and all text tokens.

Image and text tokens each go through their own small projection (linear, LayerNorm,
GELU) from Qwen's width of $2048$ down to the decoder width of $512$. We concatenate them
and add a learned positional embedding. The result is the conditioning sequence the
decoder attends to.

The decoder generates the plan by flow matching (\cref{app:generator_loss}): it starts
from random noise and refines it into a clean plan over several steps. Its input at each
step is therefore the current noisy plan, which has the same shape as the output,
$150$ frames of $36$ values. Each frame becomes one token: a linear layer maps its $36$
values to $512$, and a sinusoidal embedding encodes the frame index. Each of the $8$
blocks has self-attention across the frame tokens, cross-attention to the conditioning
sequence, and a feed-forward layer. The flow time $s$, which says how noisy the plan
still is, enters every block through adaptive LayerNorm, which scales and shifts the
features of each token. A final linear layer maps each token back to $36$ values: the
velocity that moves the noisy plan toward the clean one. This layer is initialized to
zero, so training starts from a model that predicts zero velocity everywhere. Only the
projections and the decoder are trained, $49.3$M parameters in total.

\begin{table}[h]
  \centering
  \caption{\textbf{Motion generator configuration.}}
  \label{tab:generator_hparams}
  \setlength{\tabcolsep}{4pt}
  \renewcommand{\arraystretch}{1.1}
  \footnotesize
  \begin{tabular}{@{}llc@{}}
    \toprule
    \textbf{Group} & \textbf{Parameter} & \textbf{Value} \\
    \midrule
    \multirow{4}{*}{Encoder}  & backbone              & Qwen3.5-2B (frozen) \\
                              & image resolution      & $224\times224$ \\
                              & feature layer         & last \\
                              & conditioning tokens   & $64$ image $+$ all text ($\le96$) \\
    \midrule
    \multirow{6}{*}{Decoder}  & architecture          & DiT~\cite{Peebles2022DiT} \\
                              & blocks / width / heads & $8$ / $512$ / $8$ \\
                              & FFN width             & $2048$ \\
                              & dropout               & $0.1$ \\
                              & conditioning          & cross-attention \\
                              & trainable params      & $49.3$M \\
    \midrule
    \multirow{4}{*}{Plan}     & horizon $H$           & $150$ frames ($5\,$s) \\
                              & prefix $P$            & $15$ frames ($0.5\,$s) \\
                              & frame dimension       & $36$ ($29 + 3 + 4$) \\
                              & sampler               & Euler, $15$ steps \\
    \midrule
    \multirow{8}{*}{Training} & optimizer             & AdamW, $\beta\!=\!(0.9, 0.999)$ \\
                              & learning rate         & $10^{-3}$, cosine to $10^{-5}$ \\
                              & weight decay          & $0.01$ \\
                              & gradient clipping     & $1.0$ \\
                              & batch size            & $5{,}120$ ($80\times64$ GPUs) \\
                              & epochs                & $50$ \\
                              & flow-time distribution & $\mathrm{Beta}(1.5, 1)$ \\
                              & precision             & FP32 \\
    \bottomrule
  \end{tabular}
\end{table}

\subsubsection{Training loss}
\label{app:generator_loss}
We train the decoder with flow matching~\cite{lipman2022flowmatching}. Joint angles and
pelvis translation are first normalized to zero mean and unit variance per dimension;
the quaternion is left unchanged. Given a normalized plan $x_1$ and Gaussian noise
$x_0$, we draw a flow time $s\!\sim\!\mathrm{Beta}(1.5, 1)$ and mix the two:
\begin{equation}
  x_s = s\,x_1 + (1-s)\,x_0.
\end{equation}
At $s\!=\!0$ this is pure noise and at $s\!=\!1$ it is the clean plan. The decoder
$v_\phi$, with weights $\phi$, learns to predict the velocity $x_1 - x_0$ that carries
one to the other.

For asynchronous generation, the first $P\!=\!15$ frames are frames the robot is
already committed to, so we never noise them. The decoder therefore receives two
modified inputs. The first is the plan $\tilde x_s$, whose first $P$ frames come from the
clean plan $x_1$ and whose remaining frames come from the noisy plan $x_s$. The second
is the per-frame flow time $\tilde s$, which is $1$ on the first $P$ frames, marking them
as clean, and $s$ on the remaining frames. The loss covers only the frames the decoder
has to generate:
\begin{equation}
  \mathcal{L}(\phi) = \mathbb{E}
  \left[ \frac{1}{D\,(H-P)} \sum_{\tau=P+1}^{H}
  \big\| v_\phi(\tilde x_s, \tilde s, c)_\tau - (x_1 - x_0)_\tau \big\|^2 \right],
\end{equation}
where $c$ is the conditioning sequence, $H\!=\!150$ is the plan length, $D\!=\!36$ is the number of values per frame, and the subscript $\tau$ selects frame $\tau$. The model thus learns to continue whatever prefix it is given, and consecutive plans join without a jump. We train for $50$ epochs on $31.0$M samples using $64$ B200 GPUs.

At inference, the first $P$ frames of the new plan, the prefix, are copied from the
next $P$ frames of the plan the robot is currently executing. The other $H-P$ frames of
the new plan start from noise. We take $15$ Euler steps from
$s\!=\!0$ to $s\!=\!1$. After each step, we overwrite the first $P$ frames with the prefix
again, so the new plan begins exactly where the current one continues.

\subsection{Motion tracker}
\label{app:motion_tracker}

\begin{figure*}[!t]
  \centering
  \includegraphics[width=\textwidth]{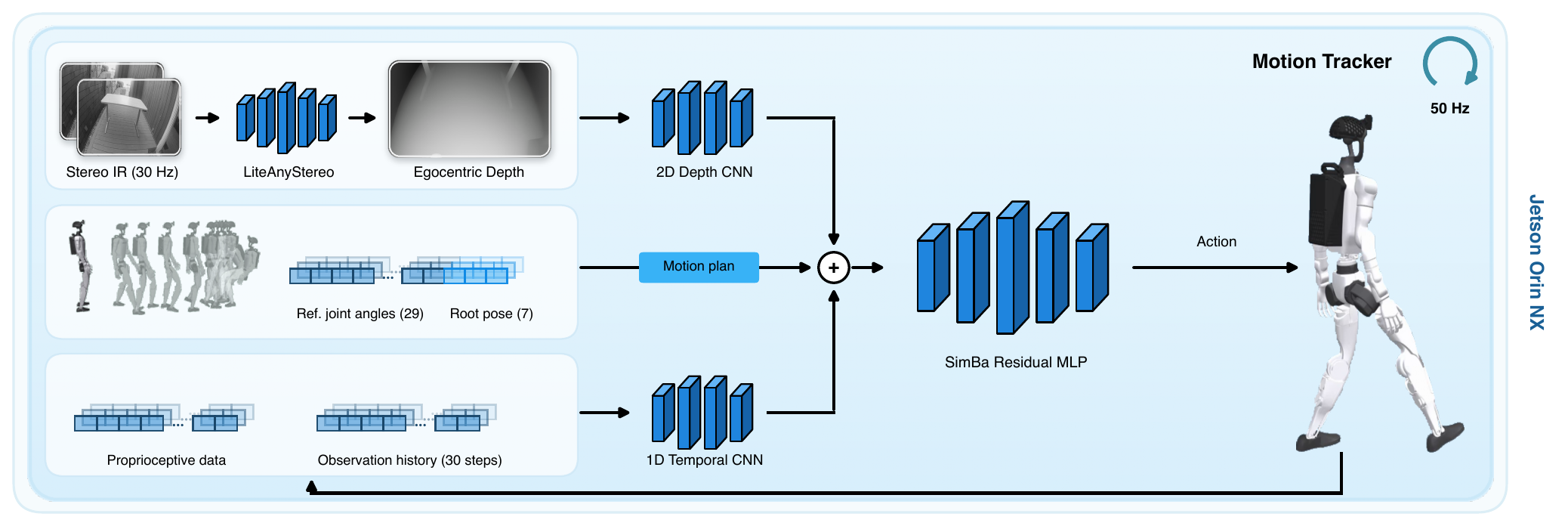}
  \caption{\textbf{Motion tracker onboard inference architecture.} On the Jetson Orin NX, the tracker runs at $50\,$Hz and fuses three onboard streams: (top left)~raw stereo IR frames from the RealSense D435i (emitter off) are mapped by LiteAnyStereo~\cite{jing2026lite} (TensorRT FP16) to a clean $48\timess 84$ egocentric depth image and encoded by a 2D CNN; (bottom left)~a $30$-step proprioceptive history together with the next $30$ plan frames is encoded by a 1D temporal CNN; and (center)~both feature vectors are concatenated with the current proprioceptive state and $5$-frame motion plan lookahead and passed through a SimBa residual MLP~\cite{lee2025simba} to output $29$ target joint positions for the low-level PD controller.}
  \label{fig:tracker_arch}
  \vspace{-6pt}
\end{figure*}

The motion tracker (\cref{sec:tracker}) is a $50\,$Hz policy that executes the kinematic motion plan from the generator on the robot (\cref{fig:tracker_arch}). On hardware, it runs in TensorRT on the robot's onboard Jetson Orin NX inside a real-time C++ process. At each step, it reads the upcoming frames of the motion plan, joint positions and velocities, IMU readings, and a depth image from the downward-facing RealSense D435i, and outputs $29$ target joint positions for the low-level PD controller (simulated at $200\,$Hz across four physics substeps per policy step, and tracked on hardware by the onboard motor controllers).

\subsubsection{Three-phase training}
Training a single depth-conditioned policy from scratch across all skills fails because contact dynamics and reward scales vary widely across tasks, and rendering depth slows early RL exploration. We therefore train the tracker in three phases in Isaac Lab~\cite{nvidia2025isaaclab} (\cref{tab:tracker_hparams}).

\emph{Phase 1: Privileged experts.}
To avoid negative transfer across disparate behaviors, we first train a separate PPO expert for each skill family: locomotion and obstacle traversal on flat ground (\textsc{Go-To}, \textsc{Door}, \textsc{Crouch}, \textsc{Crawl}, and retargeted human motion clips), object manipulation (\textsc{Push}), and object interaction (\textsc{Sit}). Each expert observes privileged simulator state, a $10$-frame reference window with extremity errors, a $30$-step history, and a local height map of the terrain and objects. Episodes start from random frames within each clip; to focus training on hard transitions, starting frames are sampled in proportion to per-segment failure rates~\cite{he2024omnih2o}.

\emph{Phase 2: Multi-teacher distillation.}
We distill the frozen experts into a single student policy that uses only onboard sensors (proprioception, IMU, the motion plan, and egocentric depth) in a shared scene containing flat ground, furniture, and movable objects. At each step $t$, simulation instance $i$ is routed to the expert $\pi^{\text{teacher}}_{k(i)}$ for its active skill $k(i)$, avoiding the blending artifacts of action averaging. Rollouts follow a hybrid behavior cloning and DAgger strategy~\cite{ross2011reduction, he2025viral} ($20\%$ teacher-driven rollouts and $80\%$ student-driven rollouts), and the student with weights $\theta$ minimizes the squared error between its mean action $\bmu_\theta(\oo^{\text{stu}}_{i,t})$ and the routed expert's mean action $\bmu^{\text{teacher}}_{k(i)}(\oo^{\text{priv}}_{i,t})$:
\begin{equation}
  \lL_{\text{distill}}(\theta) = \mathbb{E}_{i,t}\!\left[\norm{\bmu_\theta\big(\oo^{\text{stu}}_{i,t}\big) - \bmu^{\text{teacher}}_{k(i)}\big(\oo^{\text{priv}}_{i,t}\big)}^2\right].
  \label{eq:tracker_distill}
\end{equation}

\emph{Phase 3: Multi-critic fine-tuning.}
Distillation is bounded by teacher competence, and partial observability leaves a residual tracking gap~\cite{wu2026php, wang2026omnixtreme}. In Phase~3, we initialize the actor from the distilled student and train it with PPO with depth rendering in the loop across $4$ H100 GPUs ($16{,}384$ environments). Because object interaction skills include extra object rewards (\cref{tab:tracker_rewards}), their returns have a different scale than locomotion, which would cause a single shared critic to skew the policy gradient. We resolve this with multi-critic PPO: the student actor is shared across all skills, while $K$ independent privileged critics $\{V_{\psi_k}\}_{k=1}^K$, one per skill family, are initialized from the Phase~1 experts' critics. Each critic is updated only on transitions from its own skill, and the generalized advantage estimate $A_{i,t}$ is standardized separately within each skill's batch:
\begin{equation}
  \hat A_{i,t} = \frac{A_{i,t} - \mu_{k(i)}}{\sigma_{k(i)} + \eps},
  \label{eq:tracker_multicritic}
\end{equation}
where $\mu_k$ and $\sigma_k$ are the mean and standard deviation of the advantages for skill $k$. The shared actor then updates across all skills using the normalized advantages $\hat A_{i,t}$.

\begin{table}[!t]
  \centering
  \caption{\textbf{Motion tracker architecture and training hyperparameters.}}
  \label{tab:tracker_hparams}
  \setlength{\tabcolsep}{1.8pt}
  \renewcommand{\arraystretch}{1.08}
  \scriptsize
  \begin{tabular}{@{}lccc@{}}
    \toprule
    \textbf{Parameter} & \textbf{Phase 1} & \textbf{Phase 2} & \textbf{Phase 3} \\
    \midrule
    \multicolumn{4}{@{}l}{\emph{Network architecture}} \\
    History encoder (1D CNN) & $[64, 128, 128]$ & $[32, 64, 128]$ & $[32, 64, 128]$ \\
    Spatial input            & $17\timess 11$ height & $48\timess 84$ depth & $48\timess 84$ depth \\
    Spatial encoder (2D CNN) & $[32, 64, 128]$ & $[16, 32, 64, 128]$ & $[16, 32, 64, 128]$ \\
    SimBa MLP~\cite{lee2025simba} ($L\timess W$) & $3\timess 768$ & $2\timess 512$ & $2\timess 512$ \\
    Reference lookahead & $10$ ($0.2\,$s) & $5$ ($0.1\,$s) & $5$ ($0.1\,$s) \\
    \midrule
    \multicolumn{4}{@{}l}{\emph{Training}} \\
    Algorithm                & PPO & DAgger (20\% teacher) & Multi-critic PPO \\
    Parallel envs (GPUs)     & $8{,}192$ ($1$) & $4{,}096$ ($1$) & $16{,}384$ ($4$) \\
    Rollout steps per env    & $24$ & $32$ & $24$ \\
    Training iterations      & $50{,}000$ & $30{,}000$ & $50{,}000$ \\
    Learning rate (Adam)     & $10^{-3}$ & $10^{-4}$ & $3\timess 10^{-4}$ \\
    Epochs / mini-batches    & $5$ / $8$ & $5$ / $8$ & $5$ / $8$ \\
    PPO clip / entropy       & $0.2$ / $0.008$ & --- & $0.2$ / $0.001$ \\
    Discount $\gamma$ / GAE $\lambda$ & $0.99$ / $0.95$ & --- & $0.99$ / $0.95$ \\
    \bottomrule
  \end{tabular}
  \vspace{-4pt}
\end{table}

\subsubsection{Observations and network architecture}
We track $|\bB|\!=\!14$ body links: the pelvis (which carries the IMU and serves as the anchor link), the torso, and the left and right hip, knee, ankle, shoulder, elbow, and wrist links, with the two ankles and two wrists forming the extremity set $\bB_{\text{ee}}$.

\emph{Asymmetric observations.}
The Phase~1 experts and Phase~3 critics observe noiseless simulator state: base linear and angular velocity, projected gravity, joint positions and velocities, the previous action, the 3D poses of all $14$ tracked links in the pelvis frame, the pelvis pose error, a $10$-frame ($0.2\,$s) future reference window, a $30$-step ($0.6\,$s) history, and a $1.6\,\text{m}\timess 1.0\,\text{m}$ height map at $0.1\,$m resolution ($17\timess 11$). Following RPL~\cite{zhang2026rpl}, heights are measured relative to the feet so the elevation profile of furniture or objects stays constant when the pelvis lowers during crouching or sitting. The deployable student (\cref{fig:tracker_arch}) drops all world-frame velocities and link poses, observing only noisy IMU angular velocity and gravity, joint positions and velocities, the previous action, the pelvis orientation error, a shorter $5$-frame ($0.1\,$s) reference window, the next 30 plan frames, a $30$-step history, and a $48\timess 84$ depth image.

\emph{Simulated and onboard depth.}
Standard depth ray-casters in simulation ignore the robot's own body, which would let a downward-looking head camera see through the arms and legs. Following RPL~\cite{zhang2026rpl}, we render simulated depth ($25\,$Hz) with a two-pass Warp ray-caster that queries both the scene meshes and the robot's articulated link meshes, taking the closer hit at each pixel so the student observes realistic self-occlusion during manipulation, crawling, and sitting. Depth values are clipped to $[0.2, 2.5]\,$m and mapped via $\log(1+d)$. On hardware, the RealSense D435i's built-in stereo matcher produces heavy speckle noise, occlusion shadows, and invalid holes. Rather than simulating these artifacts, we disable the camera's active IR dot projector and stream the raw left and right infrared images at $30\,$Hz into LiteAnyStereo~\cite{jing2026lite} ($7.6\,$M parameters, compiled to TensorRT FP16 on the Jetson Orin NX). The predicted disparity is converted to metric depth, downsampled to $48\timess 84$ by averaging only valid measured pixels within each cell (mapping missing returns to the $2.5\,$m far plane so holes do not bias nearby obstacles), clipped to $[0.2, 2.5]\,$m, and log-transformed before being passed to the $50\,$Hz control loop through lock-free triple buffers (\cref{fig:tracker_arch}).

\emph{Policy network.}
As shown in \cref{fig:tracker_arch} and \cref{tab:tracker_hparams}, a 3-layer 1D CNN encodes the $30$-step observation history, and a 4-layer stride-$2$ 2D CNN encodes the $48\timess 84$ depth image (or a 3-layer 2D CNN over the $17\timess 11$ height map for the experts and critics). Both feature vectors are concatenated with the current state and $5$-frame motion plan and passed through a SimBa residual MLP~\cite{lee2025simba} with pre-LayerNorm feed-forward blocks (GELU) and a linear output head to predict the $29$ target joint positions.

\begin{table}[!t]
  \centering
  \caption{\textbf{Reward terms optimized in Phase 1 and Phase 3.} Pose and velocity errors are $\ee_{p,b} = \pp_b^{\text{ref}} - \pp_b$, $\ee_{R,b} = \mathrm{Log}(\RR_b^{\text{ref}}\RR_b\trans)$, $\ee_{v,b} = \vv_b^{\text{ref}} - \vv_b$, and $\ee_{\omega,b} = \bm{\omega}_b^{\text{ref}} - \bm{\omega}_b$ ($\tilde{\ee}$ denotes pelvis-aligned relative errors~\cite{liao2025beyondmimic}). Object terms are gated by hand proximity $g_{\text{hand}} = \exp(-(d_{\text{hand}}/0.3)^2)$~\cite{tessler2024maskedmimic, zhao2025resmimic}.}
  \label{tab:tracker_rewards}
  \setlength{\tabcolsep}{3.0pt}
  \renewcommand{\arraystretch}{1.08}
  \scriptsize
  \begin{tabular}{@{}llc@{}}
    \toprule
    \textbf{Reward Term} & \textbf{Expression} & \textbf{Weight} \\
    \midrule
    \multicolumn{3}{@{}l}{\emph{Motion tracking~\cite{liao2025beyondmimic}}} \\
    Pelvis position        & $\exp\!\big(-\norm{\ee_{p,\text{pelvis}}}^2 / 0.3^2\big)$ & $1.0$ \\
    Pelvis orientation     & $\exp\!\big(-\norm{\ee_{R,\text{pelvis}}}^2 / 0.4^2\big)$ & $1.0$ \\
    Link positions         & $\exp\!\big(-\frac{1}{|\bB|}\sum_{b\in\bB}\norm{\tilde{\ee}_{p,b}}^2 / 0.3^2\big)$ & $1.0$ \\
    Link orientations      & $\exp\!\big(-\frac{1}{|\bB|}\sum_{b\in\bB}\norm{\tilde{\ee}_{R,b}}^2 / 0.4^2\big)$ & $1.0$ \\
    Link linear velocities & $\exp\!\big(-\frac{1}{|\bB|}\sum_{b\in\bB}\norm{\ee_{v,b}}^2 / 0.5^2\big)$ & $1.0$ \\
    Link angular velocities & $\exp\!\big(-\frac{1}{|\bB|}\sum_{b\in\bB}\norm{\ee_{\omega,b}}^2 / 1.5^2\big)$ & $1.0$ \\
    \midrule
    \multicolumn{3}{@{}l}{\emph{Object interaction~\cite{tessler2024maskedmimic, zhao2025resmimic}}} \\
    Object position        & $g_{\text{hand}}\exp\!\big(-\norm{\ee_{p,\text{obj}}}^2 / 0.3^2\big)$ & $1.0$ \\
    Object orientation     & $g_{\text{hand}}\exp\!\big(-\norm{\ee_{R,\text{obj}}}^2 / 0.4^2\big)$ & $1.0$ \\
    Object drop            & $g_{\text{hand}}\max(0,\, z_{\text{obj}}^{\text{ref}} - z_{\text{obj}} - 0.15)$ & $-1.0$ \\
    \midrule
    \multicolumn{3}{@{}l}{\emph{Regularization and hardware safety~\cite{wang2026omnixtreme}}} \\
    Action rate            & $\norm{\aa_t - \aa_{t-1}}_2^2$ & $-0.05$ \\
    Joint position limits  & $\sum_j \max(0,\, q_j - q_j^{\max},\, q_j^{\min} - q_j)$ & $-100$ \\
    Undesired contacts     & $\sum_{b\notin\bB_{\text{ee}}} \mathbb{I}(\norm{\ff_b^{\text{contact}}} > 1\,\text{N})$ & $-0.1$ \\
    Knee braking power     & $\sum_{j\in\text{knees}} \max(0,\, -\tau_j\dot{q}_j - 150)^2 / 500^2$ & $-10.0$ \\
    Stance foot slip       & $\sum_{f\in\text{feet}} \norm{\vv_f^{xy}}\,\mathbb{I}_{\text{stance}}(f)$ & $-0.2$ \\
    \bottomrule
  \end{tabular}
  \vspace{-4pt}
\end{table}

\subsubsection{Rewards and terminations}
Phase~1 and Phase~3 optimize the reward in \cref{tab:tracker_rewards}.

\emph{Motion and object rewards.}
Six Gaussian kernels track the global pelvis pose and the relative poses and velocities of the $14$ tracked links $\bB$, with relative link targets aligned to the robot's current horizontal pelvis position and yaw so horizontal drift is not penalized twice~\cite{liao2025beyondmimic}. For object manipulation (such as pushing a box), pose imitation alone fails when the robot copies the arm motion in free air while leaving the object stationary. Following ResMimic~\cite{zhao2025resmimic} and MaskedMimic~\cite{tessler2024maskedmimic}, we measure the distance from the wrists to the object, $d_{\text{hand}} = \min_{h\in\text{wrists}} \norm{\pp_h - \pp_{\text{obj}}}$, and gate the object tracking terms by $g_{\text{hand}} = \exp(-(d_{\text{hand}}/0.3)^2)$ so the policy is rewarded for moving the object only when the hands are near the object.

\emph{Safety penalties and terminations.}
Alongside penalties on action rate, soft joint-limit violations, non-extremity contacts above $1\,$N, and foot slip during stance, we penalize excessive negative mechanical power $P_j = \tau_j \dot{q}_j < 0$ (regenerative braking) at the knees during deep crouches and sitting. Following OmniXtreme~\cite{wang2026omnixtreme}, braking power above a $150\,$W deadband is penalized to prevent policies from exploiting simulator torque clipping and tripping hardware overcurrent protection. Phase~1 terminates failed episodes using strict 3D cutoffs ($0.5\,$m pelvis, $0.3\,$m extremities $\bB_{\text{ee}}$); in Phases~2 and~3, because the student has no global position sensor and naturally drifts horizontally, we relax the 3D pelvis cutoff to $0.8\,$m and check vertical drift ($0.3\,$m pelvis, $0.375\,$m extremities) so valid obstacle traversals are not cut short~\cite{wu2026php, wang2026omnixtreme}.

\begin{table}[!t]
  \centering
  \caption{\textbf{Domain randomization ranges across training phases.}  Noise and depth rows apply to the student (Phase 2); Phase 1 experts observe noiseless state without depth.}
  \label{tab:tracker_dr}
  \setlength{\tabcolsep}{1.6pt}
  \renewcommand{\arraystretch}{1.08}
  \scriptsize
  \begin{tabular}{@{}lcc@{}}
    \toprule
    \textbf{Parameter} & \textbf{Phases 1--2} & \textbf{Phase 3} \\
    \midrule
    \multicolumn{3}{@{}l}{\emph{Actuators and robot inertia~\cite{liao2025beyondmimic, wang2026omnixtreme}}} \\
    PD stiffness / damping scale            & $1.0$ / $1.0$            & $[0.85, 1.15]$ / $[0.75, 1.3]$ \\
    Motor armature scale                    & $1.0$                    & $[0.90, 1.20]$ \\
    Joint torque limit scale                & $1.0$                    & $[0.80, 1.05]$ \\
    Default joint offset (rad)              & $\pm 0.01$               & $\pm 0.01$ \\
    Command latency (ms)                    & $[5, 25]$                & $[15, 40]$ \\
    Torso CoM $\Delta x$ / $\Delta y,z$ (m) & $\pm 0.025$ / $\pm 0.05$ & $\pm 0.025$ / $\pm 0.05$ \\
    Limb mass scale                         & $1.0$                    & $[0.90, 1.10]$ \\
    External velocity pushes (m/s)          & $\pm 0.50$               & $\pm 0.50$ \\
    \midrule
    \multicolumn{3}{@{}l}{\emph{Ground surface and objects}} \\
    Ground friction (static / dynamic)      & $[0.3, 1.6]$ / $[0.3, 1.2]$ & $[0.3, 1.6]$ / $[0.3, 1.2]$ \\
    Ground restitution                      & $[0.0, 0.4]$             & $[0.0, 0.5]$ \\
    Ground stiffness (kN/m) / $\zeta$       & rigid                    & $[1.3, 6.5]$ / $[0.3, 1.0]$ \\
    Object added mass (kg) / friction       & $[0.0, 0.5]$ / $[0.6, 0.8]$ & $[0.0, 0.5]$ / $[0.6, 0.8]$ \\
    \midrule
    \multicolumn{3}{@{}l}{\emph{Proprioception and depth camera~\cite{zhang2026rpl, wu2026php}}} \\
    IMU $\bm{\omega}_b$ (rad/s) / gravity noise & $\pm 0.20$ / $\pm 0.05$  & $\pm 0.20$ / $\pm 0.05$ \\
    Joint $\qq$ (rad) / $\dot{\qq}$ (rad/s) noise & $\pm 0.01$ / $\pm 0.50$ & $\pm 0.01$ / $\pm 0.50$ \\
    Camera mount offset                     & $\pm 2.5\,$cm, $\pm 2.5^\circ$ & $\pm 2.5\,$cm, $\pm 2.5^\circ$ \\
    Stereo speckle $\sigma(d)$ / bias (m)   & $0.001\,d^2$ / $\pm 0.01$ & $0.001\,d^2$ / $\pm 0.01$ \\
    Pixel dropout / frame delay             & $3\%$ / $1\text{--}3$    & $3\%$ / $1\text{--}3$ \\
    Occluders / side masking~\cite{zhang2026rpl} & $0\text{--}2$ / $\le\!12$ cols & $0\text{--}2$ / $\le\!12$ cols \\
    \bottomrule
  \end{tabular}
  \vspace{-4pt}
\end{table}

\subsubsection{Actuator modeling and domain randomization}
To minimize the sim-to-real gap, we combine system-identified actuator modeling with a two-tier domain randomization schedule (\cref{tab:tracker_dr}).

\emph{System identification and PD control.}
Following BeyondMimic~\cite{liao2025beyondmimic} and OmniXtreme~\cite{wang2026omnixtreme}, we derive nominal joint stiffness $k_{p,j}$, damping $k_{d,j}$, and torque-normalized action scale $\alpha_j$ from each motor's rotor inertia $I_j$ using a natural frequency $\omega_n\!=\!10\,$Hz ($20\pi\,$rad/s) and damping ratio $\zeta\!=\!2.0$:
\begin{equation}
  k_{p,j} = I_j \omega_n^2, \;\; k_{d,j} = 2\zeta I_j \omega_n, \;\; \alpha_j = \frac{0.25\,\tau_{\max,j}}{k_{p,j}},
  \label{eq:pd_gains}
\end{equation}
where $\tau_{\max,j}$ is the joint torque limit, and the policy commands target positions $\qq_t^{\text{tar}} = \qq^0 + \bm{\alpha}\odot\aa_t$. To match our physical Unitree G1 ($37.56\,$kg with the rear Mac mini enclosure and head camera mount), we replay open-loop trajectories on the robot and fit the effective per-joint stiffness, damping, rotor armature, torque limits, and transport delay to the measured response. All training phases use this identified model with a delayed PD actuator.

\emph{Physical, ground, and depth randomization.}
Randomization is kept narrow in Phases~1 and~2 to preserve teacher fidelity, then widened in Phase~3 (\cref{tab:tracker_dr}). Phase~3 randomizes PD gains, rotor armatures, torque limits, link masses, and randomized command latency ($15\text{--}40\,$ms) around the identified hardware model. We also randomize object mass and friction, as well as ground friction and contact compliance (stiffness $1.3\text{--}6.5\,$kN/m and damping ratio $\zeta \in [0.3, 1.0]$) so the policy transfers across both rigid floors and compliant carpets. Simulated depth frames in Phases~2 and~3 are corrupted with camera-mount jitter ($\pm 2.5\,$cm, $\pm 2.5^\circ$), random rectangular occluders, distance-dependent stereo noise ($\sigma(d)\!=\!0.001\,d^2$), $3\%$ pixel dropout to the far plane, $1\text{--}3$ frames of latency, and side masking~\cite{zhang2026rpl} of up to $12$ peripheral columns.

\subsection{Orchestrator}
\label{app:orchestrator}

\subsubsection{Model} The orchestrator runs on the Mac mini using Qwen3.6-35B-A3B (4-bit,
MLX). It shares one serving engine with the motion generator
(\cref{app:system_software}), so planning needs no network connection. It can also
call frontier closed-source models through their API, which may
give better planning performance.

\subsubsection{Subtask descriptions} The orchestrator learns what the motion generator can do
from a description of each subtask in its prompt: when the subtask applies and what its
completion looks like. When a subtask is selected, the generator receives the subtask's
text prompt. \cref{tab:orch_skills} lists the subtasks and their generator prompts; the
full descriptions are in the plan-authoring prompt below.

\subsubsection{Plan decomposition} When an instruction arrives, the orchestrator looks at the
current egocentric frame and breaks the instruction into an ordered plan of up to eight
subtasks from \cref{tab:orch_skills}, together with a short reply to the operator. The output
follows a fixed JSON schema, so every step names a subtask the robot has. Each call sends a fixed
system prompt, which the serving engine caches, and a user message with the operator's
request and the newest egocentric frame. An example is shown below.

\par\smallskip\noindent\rule{\linewidth}{0.4pt}\par
{\scriptsize\ttfamily\raggedright\sloppy\setlength{\parskip}{1pt}
\hangindent=1em\hangafter=1 SYSTEM PROMPT:\par
\hangindent=1em\hangafter=1 You are a humanoid robot that performs whole-{}body motion subtasks. Understand the request, plan an ordered task list of subtasks, execute it, and reply to the human in one short sentence (no numbers or measurements). If the request needs no action, just reply. Never report results you did not observe.\par
\smallskip
\hangindent=1em\hangafter=1 SUBTASKS:\par
\hangindent=2em\hangafter=1\hspace*{1em}go\_to\_chair() -{} walk to a chair across clear floor and stop about a metre in front of it. Use before sit.\par
\hangindent=4em\hangafter=1\hspace*{3em}done when: the chair is close and in view.\par
\hangindent=2em\hangafter=1\hspace*{1em}go\_to\_box() -{} walk to a box across clear floor and stop in front of it.\par
\hangindent=4em\hangafter=1\hspace*{3em}done when: the box is close and in view.\par
\hangindent=2em\hangafter=1\hspace*{1em}go\_to\_human() -{} walk to a person across clear floor and stop.\par
\hangindent=4em\hangafter=1\hspace*{3em}done when: the person is close and in view.\par
\hangindent=2em\hangafter=1\hspace*{1em}follow() -{} stay with a target that can move away, such as a person leading the way.\par
\hangindent=4em\hangafter=1\hspace*{3em}done when: never on your own judgment; the operator ends it.\par
\hangindent=2em\hangafter=1\hspace*{1em}sit() -{} sit down on a seat directly ahead with a clear path.\par
\hangindent=4em\hangafter=1\hspace*{3em}done when: the robot is seated.\par
\hangindent=2em\hangafter=1\hspace*{1em}crouch\_under() -{} pass under an obstacle by bending the knees. The default for getting under a table.\par
\hangindent=4em\hangafter=1\hspace*{3em}done when: the robot is out the far side and upright.\par
\hangindent=2em\hangafter=1\hspace*{1em}crawl\_through() -{} crawl under an obstacle too low to crouch under.\par
\hangindent=4em\hangafter=1\hspace*{3em}done when: the obstacle is behind the robot.\par
\hangindent=2em\hangafter=1\hspace*{1em}sidewalk() -{} side-{}step through a half-{}open door or a tight gap.\par
\hangindent=4em\hangafter=1\hspace*{3em}done when: the robot is through the gap.\par
\hangindent=2em\hangafter=1\hspace*{1em}push\_box() -{} push a box directly in front of the robot toward the goal.\par
\hangindent=4em\hangafter=1\hspace*{3em}done when: the box has visibly moved.\par
\hangindent=2em\hangafter=1\hspace*{1em}stand() -{} return to the standing pose and wait.\par
\hangindent=4em\hangafter=1\hspace*{3em}done when: the robot is upright and still.\par
\hangindent=2em\hangafter=1\hspace*{1em}hold() -{} keep the current pose; nothing to act on yet.\par
\hangindent=4em\hangafter=1\hspace*{3em}done when: what was awaited has arrived.\par
\smallskip
\hangindent=1em\hangafter=1 USER MESSAGE:\par
\hangindent=2em\hangafter=1\hspace*{1em}Crouch under the table, then sit on the chair.\par
\hangindent=2em\hangafter=1\hspace*{1em}ego frame: now\par
\hangindent=2em\hangafter=1\hspace*{1em}\textless{}current egocentric image\textgreater{}\par
}\par\noindent\rule{\linewidth}{0.4pt}\par\smallskip

\subsubsection{Transitions} While a step runs, the orchestrator makes step decisions back to
back, at its maximum rate. At each decision it looks at the current frame and chooses one
action: the step is done (\texttt{complete\_step}), the whole task is done
(\texttt{finish\_mission}), retry the current step (\texttt{retry\_step}), insert a subtask to
clear a blocked step (\texttt{insert\_step}), rewrite the plan (\texttt{set\_plan}), or ask
the operator (\texttt{ask\_operator}). An example is shown below.

\par\smallskip\noindent\rule{\linewidth}{0.4pt}\par
{\scriptsize\ttfamily\raggedright\sloppy\setlength{\parskip}{1pt}
\hangindent=1em\hangafter=1 SYSTEM PROMPT:\par
\hangindent=1em\hangafter=1 You are a humanoid robot that performs whole-{}body motion subtasks. Your responsibility is to supervise the plan while it runs: at each check, look at the current image and decide the next action by calling ONE tool:\par
\hangindent=2em\hangafter=1\hspace*{1em}set\_plan(steps) -{} write or rewrite the plan.\par
\hangindent=2em\hangafter=1\hspace*{1em}complete\_step(success) -{} the current step is finished; advance.\par
\hangindent=2em\hangafter=1\hspace*{1em}retry\_step() -{} the step stopped early and its goal is still ahead; run it again.\par
\hangindent=2em\hangafter=1\hspace*{1em}insert\_step(subtask) -{} the step is blocked; run another subtask first, then resume.\par
\hangindent=2em\hangafter=1\hspace*{1em}ask\_operator(question) -{} you are unsure or lost sight of the target.\par
\hangindent=2em\hangafter=1\hspace*{1em}finish\_mission() -{} the whole task is done.\par
\smallskip
\hangindent=1em\hangafter=1 GENERAL RULES:\par
\hangindent=2em\hangafter=1\hspace*{1em}-{} Judge each step from the current image, with the current pelvis height as support.\par
\hangindent=2em\hangafter=1\hspace*{1em}-{} Prefer retry\_step while the goal is still ahead and reachable, insert\_step when a different motion is needed to clear the way, and ask\_operator rather than guessing when the scene is ambiguous.\par
\hangindent=2em\hangafter=1\hspace*{1em}-{} A step toward a static object is done once the robot has arrived and stopped.\par
\hangindent=2em\hangafter=1\hspace*{1em}-{} A step that follows a moving target is never done on your own judgment: keep following while the target is visible, ask the operator if you lose sight of it, and let only the operator end it.\par
\hangindent=2em\hangafter=1\hspace*{1em}-{} Complete a step only when the robot has stopped and is back in a stable pose; while it is still moving or low, only observe, insert a step, or ask.\par
\smallskip
\hangindent=1em\hangafter=1 USER MESSAGE:\par
\hangindent=2em\hangafter=1\hspace*{1em}Task: Crouch under the table, then sit on the chair.\par
\hangindent=2em\hangafter=1\hspace*{1em}Plan:\par
\hangindent=3em\hangafter=1\hspace*{2em}1. [completed] crouch\_under -{} crouch under the table\par
\hangindent=3em\hangafter=1\hspace*{2em}2. [in\_progress] go\_to\_chair -{} walk to the chair\par
\hangindent=3em\hangafter=1\hspace*{2em}3. [pending] sit -{} sit down on the chair\par
\hangindent=2em\hangafter=1\hspace*{1em}NEXT step: \textquotesingle{}sit\textquotesingle{} (acts on the chair). If the robot is already in a good position to start it (the chair is close and in view), call complete\_step.\par
\hangindent=2em\hangafter=1\hspace*{1em}This step goes to a STATIC object: once the robot has ARRIVED and stopped, the step is done -{} call complete\_step to advance to the next step.\par
\hangindent=2em\hangafter=1\hspace*{1em}Robot pose (raw): base\_height=0.78m, yaw=-165deg\par
\hangindent=2em\hangafter=1\hspace*{1em}ego frame: now\par
\hangindent=2em\hangafter=1\hspace*{1em}\textless{}current egocentric image\textgreater{}\par
}\par\noindent\rule{\linewidth}{0.4pt}\par\smallskip

\subsection{System}
\label{app:system}

\subsubsection{Software}
\label{app:system_software}
The tracker runs as a C++ process on the Jetson; the generator and
the orchestrator run as separate processes on the Mac mini. An optional
web GUI shows the robot state, the current plan, and the orchestrator's decisions. The
same processes run in MuJoCo and on the robot; only the tracker binary and the camera
source change. To enforce strict real-time determinism, the tracker is implemented in C++ on a \texttt{PREEMPT\_RT} kernel, pinned to isolated CPU cores under \texttt{SCHED\_FIFO} priority with zero dynamic heap allocation, and coupled to the motor interface via lock-free triple buffers that prevent blocking across rate boundaries.

\begin{figure}[t]
  \centering
  \resizebox{\columnwidth}{!}{%
  \begin{tikzpicture}[
      font=\small,
      node/.style={rectangle, rounded corners=4mm, draw, thick, fill=blue!8, minimum width=2.3cm, minimum height=0.8cm},
      topic/.style={rectangle, draw, fill=gray!8, font=\footnotesize\ttfamily, minimum height=0.5cm, inner sep=2.5pt},
      srv/.style={rectangle, draw, dashed, fill=orange!8, font=\footnotesize\ttfamily, minimum height=0.5cm, inner sep=2.5pt},
      pub/.style={-{Latex[length=1.8mm]}, thick},
      req/.style={-{Latex[length=1.8mm]}, thick, dashed},
    ]
    \node[node] (orch) at (0,0)      {orchestrator};
    \node[node] (gen)  at (0,-3.0)   {generator};
    \node[node] (trk)  at (0,-6.0)   {tracker};
    \node[node] (gui)  at (-4.4,-3.0) {web GUI};
    \node[topic] (plan)    at (-4.4,-1.5)  {/g1/plan};
    \node[topic] (mission) at (-2.3,-1.5)  {/g1/subtask};
    \node[topic] (obs)     at (0,-1.5)     {/g1/obs};
    \node[topic] (status)  at (2.3,-1.5)   {/g1/gen\_status};
    \draw[pub] ([yshift=1.5mm]orch.west) -| (plan.north);
    \draw[pub] (plan.south) -- (gui.north);
    \draw[pub] ([yshift=-1.5mm]orch.west) -| (mission.north);
    \draw[pub] (mission.south) |- ([yshift=1.5mm]gen.west);
    \draw[pub] ([xshift=-3mm]mission.south) |- ([yshift=1.5mm]gui.east);
    \draw[pub] (gen.north) -- (obs.south);
    \draw[pub] (obs.north) -- (orch.south);
    \draw[pub] ([xshift=-3mm]obs.north) |- (-3.55,-0.8) -- (-3.55,-2.6);
    \draw[pub] (gen.east) -| (status.south);
    \draw[pub] (status.north) |- (orch.east);
    \node[srv] (gs) at (-2.5,-4.5) {/g1/get\_state};
    \node[srv] (pp) at (0,-4.5)    {/g1/put\_plan};
    \node[srv] (gt) at (2.5,-4.5)  {/g1/get\_trim};
    \draw[req] ([xshift=-6mm]gen.south) -- ++(0,-0.45) -| (gs.north);
    \draw[req] (gen.south) -- (pp.north);
    \draw[req] ([xshift=6mm]gen.south) -- ++(0,-0.45) -| (gt.north);
    \draw[req] (gs.south) |- ([yshift=1.5mm]trk.west);
    \draw[req] (pp.south) -- (trk.north);
    \draw[req] (gt.south) |- ([yshift=1.5mm]trk.east);
    \node[topic] (state) at (0,-7.3) {/g1/state (50\,Hz)};
    \draw[pub] (trk.south) -- (state.north);
    \draw[pub] (state.west) -| (gui.south);
    \draw[pub] (state.east) -- ++(2.4,0) |- ([yshift=-1.5mm]gen.east);
  \end{tikzpicture}}
  \caption{\textbf{Communication graph.}
  The generator pulls the robot
  state (\texttt{get\_state}) and the not-yet-executed reference (\texttt{get\_trim})
  from the tracker and sends each new motion plan (\texttt{put\_plan}). The
  orchestrator sends the current subtask to the generator (\texttt{subtask}) and receives
  the egocentric frame (\texttt{obs}) and the generator's motion status
  (\texttt{gen\_status}). The web GUI subscribes to \texttt{state}, \texttt{obs},
  \texttt{subtask}, and \texttt{plan}.}
  \label{fig:channels}
\end{figure}
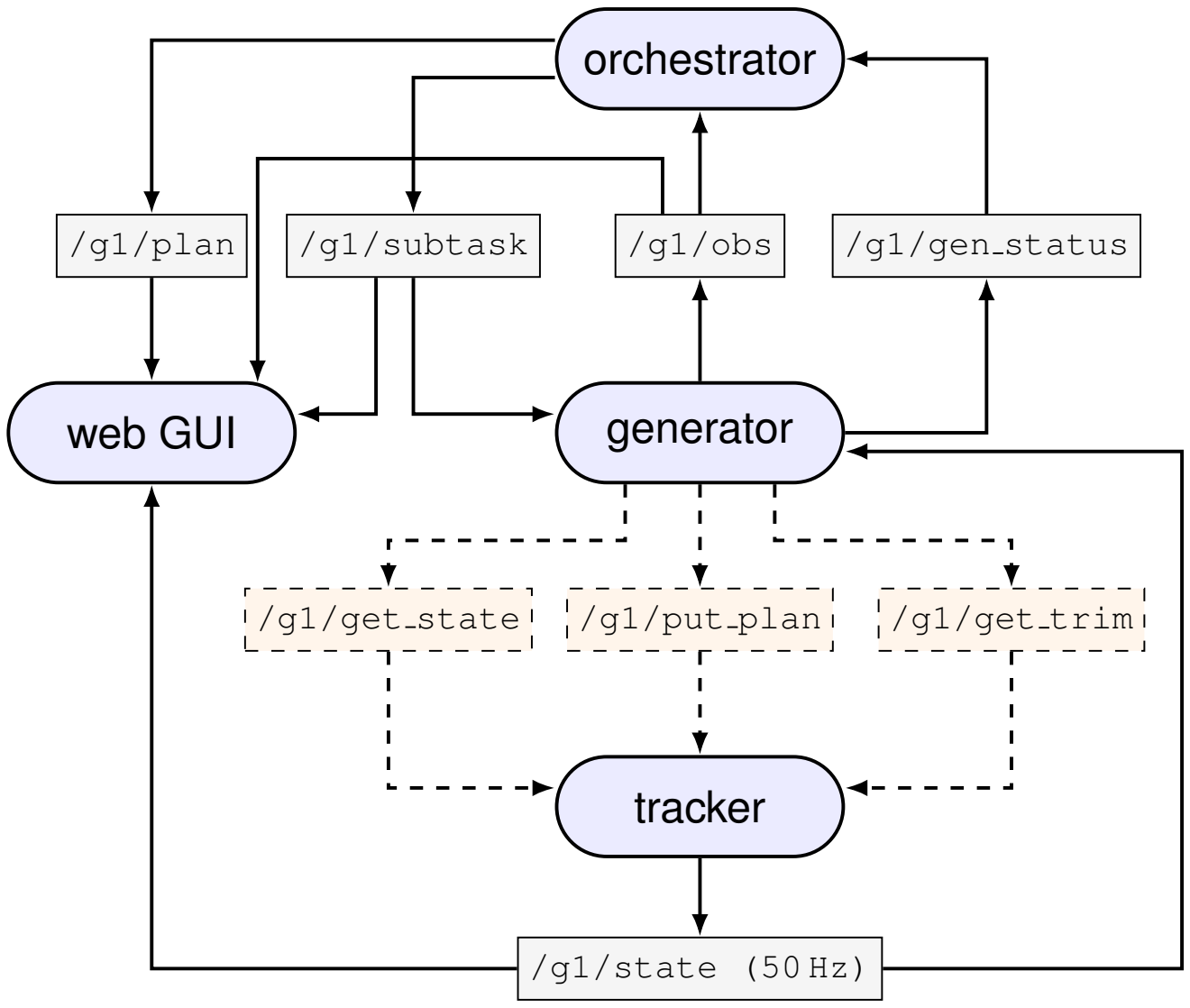

Processes talk through \texttt{halo\_comm}, our lightweight
middleware built on ZeroMQ and protobuf, with matching Python and C++ versions. It
supports request/reply services and publish/subscribe topics. Subscribers keep only
the newest message, so a slow process always sees the latest state rather than a
backlog. A registry on the robot's Jetson lets processes on different machines find
each other by channel name. \cref{fig:channels} shows the processes and channels.

\emph{Serving engine.} The generator and the local orchestrator share the Mac's GPU
through one scheduler that runs all GPU work on a single thread. Motion plans have
priority (\SI{333}{ms} deadline) and can pause an orchestrator call between model
layers, but the orchestrator keeps at least 60\% of GPU time so it is never starved.
The fixed part of the orchestrator prompt is cached, so each call only processes the
new part. The generator's DiT and Qwen3.5-2B encoder run in MLX.

\emph{Latency.} \cref{tab:latency} summarizes the measured latency of the onboard
stack on the Mac mini M4 Pro.

\begin{table}[H]
  \centering
  \caption{\textbf{Onboard latency on the Mac mini M4 Pro.} Motion generator latency is
  measured only for the model inference. Orchestrator latency is measured per transition decision, with the
  fixed prompt prefix cached.}
  \label{tab:latency}
  \footnotesize
  \begin{tabular}{@{}lc@{}}
    \toprule
    \textbf{Stage} & \textbf{Latency} \\
    \midrule
    Motion generator, alone                           & $\sim$\SI{152}{ms} \\
    Motion generator, co-served with orchestrator transition    & $\sim$\SI{155}{ms} \\
    Orchestrator transition, alone                 & $\sim$\SI{0.72}{s} \\
    Orchestrator transition, co-served with generator & $\sim$\SI{1.46}{s} \\
    \bottomrule
  \end{tabular}
\end{table}

\subsubsection{Hardware}
\label{app:system_hardware}
\begin{figure}[!tb]
    \centering
    \includegraphics[width=1\linewidth]{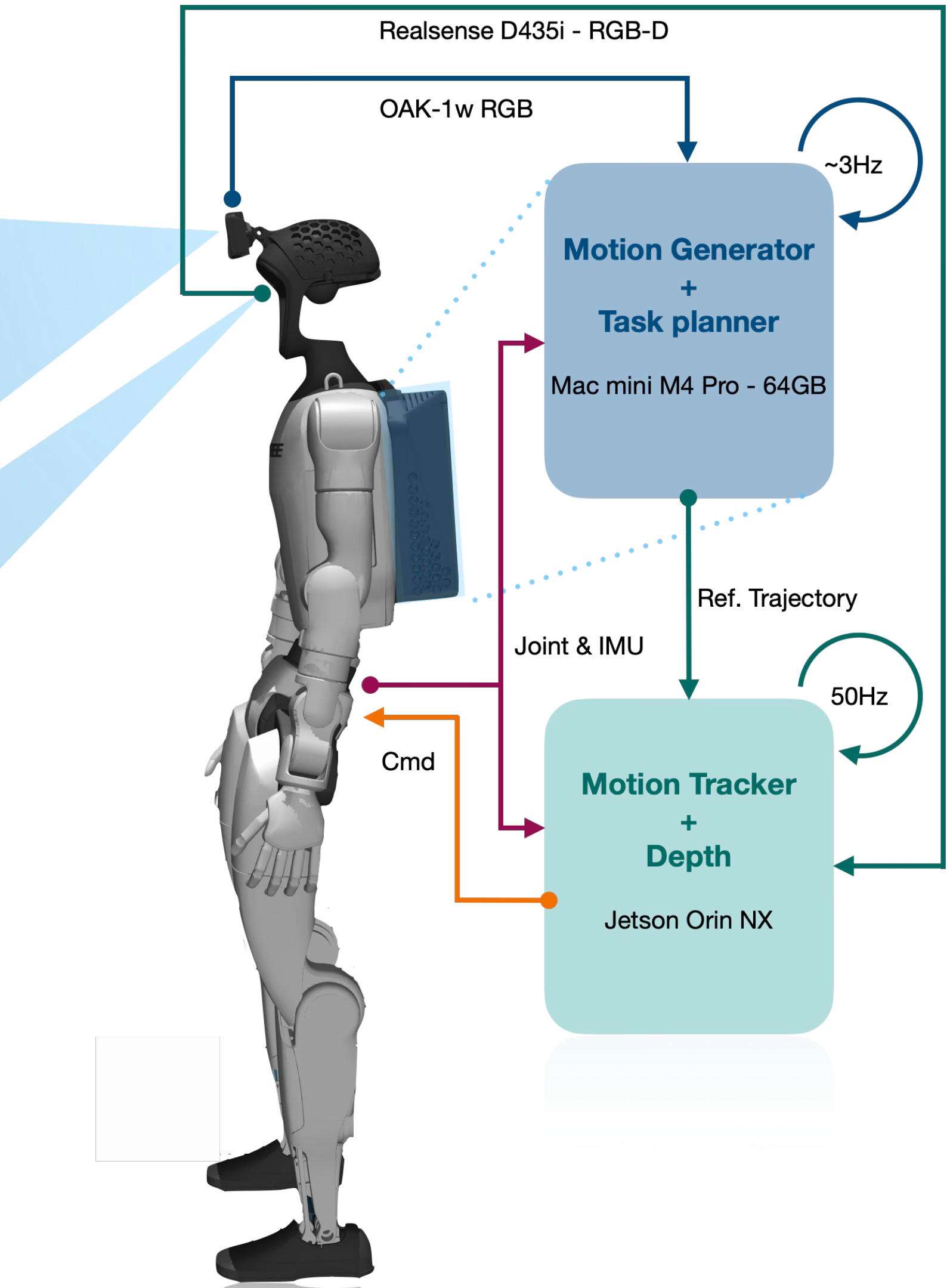}
    \caption{\textbf{Onboard compute and sensing.} A Mac~mini M4~Pro in the rear
    enclosure runs the orchestrator ($\sim\!\SI{0.5}{Hz}$) and the motion generator
    ($\sim\!\SI{3}{Hz}$). The robot's Jetson Orin NX runs the motion tracker in a
    \SI{50}{Hz} real-time loop. The OAK-1W supplies the $120^\circ$ RGB the generator
    reasons over; the RealSense~D435i supplies the $55^\circ$ depth the tracker reacts
    to. All modules run onboard, with external network calls used only when querying an optional cloud VLM for orchestration.}
    \label{fig:hw}
\end{figure}
Our hardware platform (Fig. \ref{fig:hw}) is a Unitree G1 humanoid modified for fully untethered operation with two physical additions: a rigid head mount carrying a forward-facing OAK-1W wide-angle camera, complementing the factory downward-facing RealSense D435i, and a rear protective enclosure housing a Mac mini M4 Pro (64\,GB unified memory). Powered directly from the robot's internal power bus and bridged to the factory Jetson Orin NX via gigabit Ethernet, this dual-compute architecture isolates high-capacity planning from time-critical motor control in a two-rate asynchronous cascade over ZeroMQ. In the outer loop, the OAK-1W provides a $120^\circ$ RGB stream ($1280{\times}80$, 30\,Hz) to the Mac mini, where the orchestrator and motion generator synthesize kinematic reference trajectories onboard, with off-robot communication restricted to  optional cloud VLM queries. In the inner loop on the Orin NX, the controller tracks this trajectory while ingesting near-field geometry from the factory $55^\circ$ downward-facing Intel RealSense D435i ($424{\times}240$, 30\,Hz). We bypass the D435i's noisy onboard depth pipeline altogether: we disable the active IR projector and route the raw stereo IR pair directly to a TensorRT-accelerated LiteAnyStereo~\cite{jing2026lite} model co-resident on the Orin, recovering clean depth at 30\,Hz without occlusion shadows or speckle noise and raising real sensor fidelity to match the simulator.

\begin{table}[t]
  \centering
  \caption{\textbf{Subtasks available to the orchestrator.} \emph{Generator prompt} is the
  text sent to the motion generator when the subtask is selected; the full subtask descriptions
  are in the orchestrator's system prompt (\cref{app:orchestrator}). \texttt{hold}
  and \texttt{stand} do not run the generator model: the system sends a standing or
  pose-holding reference built from the tracker's current reference.}
  \label{tab:orch_skills}
  \footnotesize
  \renewcommand{\arraystretch}{1.15}
  \begin{tabular}{@{}ll@{}}
    \toprule
    \textbf{Subtask} & \textbf{Generator prompt} \\
    \midrule
    \texttt{go\_to\_chair}   & \textit{walk to the chair} \\
    \texttt{go\_to\_box}     & \textit{walk to the box} \\
    \texttt{go\_to\_human}   & \textit{walk to the person} \\
    \texttt{follow}          & \textit{walk to the person} \\
    \texttt{sit}             & \textit{sit down on the chair} \\
    \texttt{crouch\_under}   & \textit{crouch under the table} \\
    \texttt{crawl\_through}  & \textit{crawl under the desk} \\
    \texttt{sidewalk}        & \textit{side-step through the narrow gap} \\
    \texttt{push\_box}       & \textit{push the box} \\
    \texttt{stand}           & ---  \\
    \texttt{hold}            & ---  \\
    \bottomrule
  \end{tabular}
\end{table}

\subsection{Experimental details}
\label{app:experimental_details}

\subsubsection{Metric for evaluation in simulation}
\label{app:eval_metric}
We evaluate in MuJoCo with the full deployment stack. The generator sees the head-camera view rendered with the
OAK-1W fisheye model. Simulation runs in real time, so planning latency is included.

\emph{Success criteria.} A rollout fails if the robot falls at any point: pelvis
tilt above \SI{60}{\degree} or pelvis height below \SI{0.30}{m} for at least
\SI{1}{s}. Otherwise it succeeds if its task's criteria hold.
\begin{itemize}
  \item \textsc{Crouch} and \textsc{Crawl}: the robot passes under the table (not
  around it) without colliding with it, then stands up and stops beyond it.
  \item \textsc{Door}: the robot passes through the doorway into the room without
  colliding with or crossing the door leaf, then stands still inside, clear of the door's
  swing.
  \item \textsc{Sit}: the robot ends seated within \SI{0.9}{m} of the chair (pelvis at
  \SIrange{0.35}{0.68}{m}, torso within \SI{40}{\degree} of upright).
  \item \textsc{Push}: the box moves at least \SI{0.3}{m} during the push.
  \item \textsc{Go-To}: the robot stops within \SI{1.0}{m} of the named object.
  \item Multi-task sequences: every subtask succeeds, in order.
\end{itemize}

\subsubsection{Real-world rollouts}
\label{app:real_world_rollouts}
\begin{figure*}[!t]
  \centering
  \includegraphics[width=\textwidth]{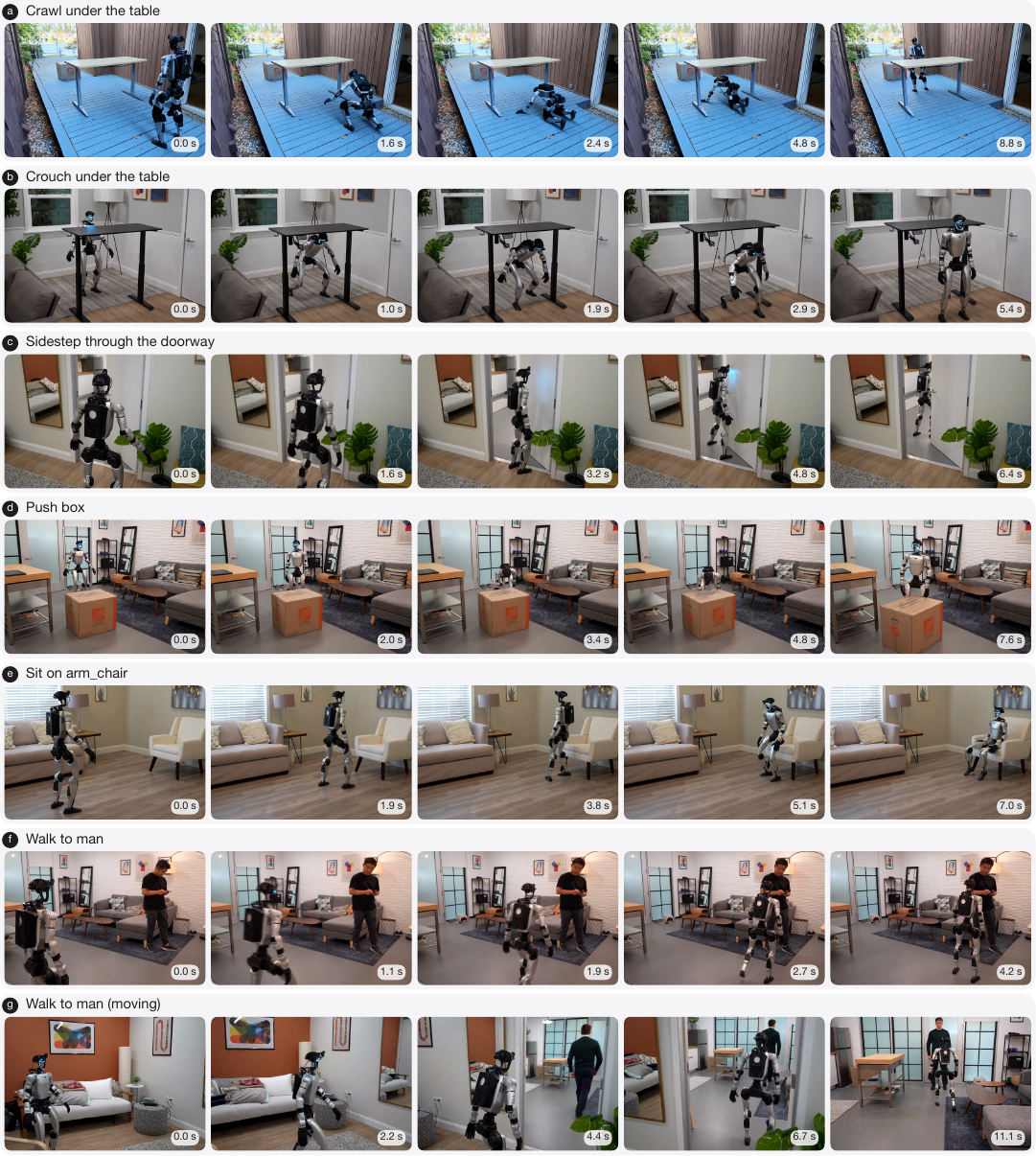}
  \caption{\textbf{Real-world rollouts, one per skill.} Each row is one single-task run on
  the physical robot, filmed by an external camera. The row header is
  the subtask prompt given to the motion generator. The five frames are spaced through
  the run, and the time in each frame is measured from the first frame of its row.}
  \label{fig:app-real-tasks}
  \vspace{-8pt}
\end{figure*}

\begin{figure*}[!t]
  \centering
  \includegraphics[width=\textwidth]{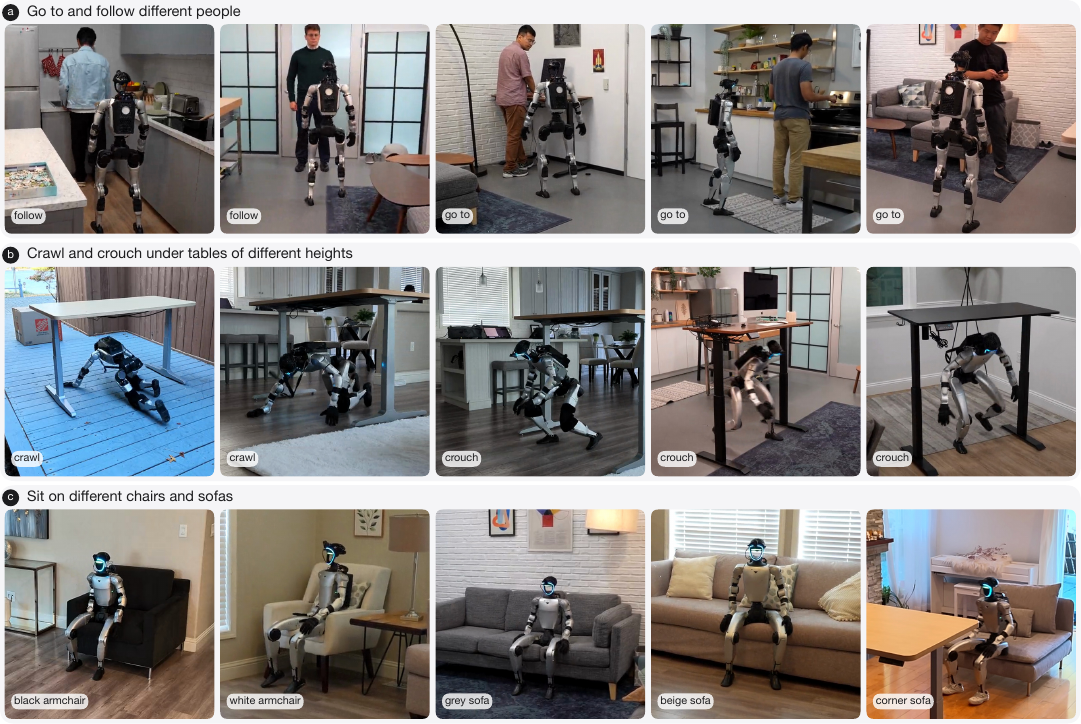}
  \caption{\textbf{Real-world generalization across people, tables, and seats.} Each tile
  is one frame from a separate real-world run.
  (a)~\textsc{Go-To} toward different people in different rooms.
  (b)~\textsc{Crawl} under two tables (outdoors and indoors) and \textsc{Crouch} under three
  desks of different heights.
  (c)~\textsc{Sit} on a black armchair, a white armchair, two sofas and a corner sofa.}
  \label{fig:app-real-generalization}
  \vspace{-8pt}
\end{figure*}

\cref{fig:app-real-tasks} shows one real-world rollout per skill, and
\cref{fig:app-real-generalization} shows the same skills in other rooms, with other people
and with other objects. All runs use the hardware of \cref{app:system_hardware}.

\subsection{Discussion}
\label{app:Discussion}
\subsubsection{Frozen vs.\ fine-tuned VLM}
The motion generator in \cref{app:motion_generator} keeps the VLM encoder frozen. We
also tried fine-tuning it, updating both Qwen's vision encoder and its language model (learning
rate $10^{-5}$ for the VLM, $10^{-3}$ for the decoder). We did not study this
systematically; the observations below come from training these variants and deploying
them on the robot.

\emph{Responsiveness.} With a fine-tuned
encoder, the robot reacts faster and moves more smoothly, most visibly when following a
person. Our explanation is that a frozen VLM knows \emph{what} is in the image but not
precisely \emph{where}. For example, a person at a bearing of $39^\circ$ and one at
$40^\circ$ produce almost the same features, so the generator cannot respond to small
movements. Fine-tuning teaches the encoder to keep this fine spatial detail.

\emph{Generalization.} Our training
data for following a person contained only one human model. With a frozen encoder, the
robot still followed many different people in the real world, because the VLM already
understands ``person'' in general. After fine-tuning, the robot only recognized people
who looked like that one model. To fix this, we re-rendered these episodes with many
different human models, created with the 3D asset generator
Hunyuan3D~\cite{hunyuan3d2025hunyuan3d}. With varied people in training, the fine-tuned model
generalizes again.

\emph{Instruction following.} In our data, the image often already reveals the task: if a chair is in view,
the command is usually ``\textit{sit on the chair}''. A fine-tuned model learns this
shortcut, relying on the image and ignoring the command. To fix this, we replace the
image with a black frame on $30\%$ of training samples. On those samples the model can
only succeed by reading the command, so it learns to follow the language command again.

\subsubsection{Motion tracker design and deployment}
\label{app:tracker_discussion}
Deploying the motion tracker across different homes and outdoor scenes exposed several physical gaps that a standard simulator with datasheet parameters and rigid floors does not capture. The observations below come from comparing these design choices on the robot.

\emph{System identification.}
Off-the-shelf simulation models of the G1 omit the internal battery, underestimate hip rotor inertia, and assume ideal, delay-free PD drives. On hardware, the physical motors deliver only $78\text{--}90\%$ of the commanded stiffness with a $25\text{--}30\,$ms end-to-end loop delay, and the ankle and waist parallel linkages exhibit higher effective inertia and friction than the serial joints. Fitting these parameters from open-loop hardware trajectories cuts joint tracking error on held-out robot motions by $62\%$ ($26.9\%$ to $10.1\%$) and reduces short-horizon prediction error during walking by about a third. More importantly, centering training on the identified robot lets Phase~3 domain randomization use narrow, realistic ranges around the true dynamics instead of wide bands that make the policy overly cautious.

\emph{Floor and carpet randomization.}
Our real-world evaluations span wooden and glossy laminate floors, tile, indoor and outdoor cement, doorway thresholds, and carpets of varying thickness, often within a single multi-task rollout. A policy trained only on rigid ground transfers well to hard surfaces (wood, tile, and cement), but struggles on plush carpet: the robot's weight sinks the feet by $1\text{--}2\,$cm into the pile, which a rigid-floor policy misinterprets as a tilt disturbance and tries to correct by continuously shuffling its feet or catching a toe when turning in place. Conversely, without low-friction randomization, the feet slip during deceleration and turns on dusty wood or smooth laminate. Randomizing both surface friction and ground contact compliance (stiffness and damping), together with the Phase~3 stance foot-slip penalty, teaches the robot to stand quietly on thick carpets while keeping firm footing on slippery hard floors.

\emph{Action smoothness and latency.}
Across trained checkpoints, raw simulation reward correlates weakly with real-world stability, whereas the action-rate smoothness penalty strongly predicts clean hardware transfer. Without sufficient action-rate regularization and randomized command latency ($15\text{--}40\,$ms), high-frequency policy actions interact with hardware transport delay and joint stiction, causing audible vibration at the ankles and wrists.

\end{document}